\pdfoutput=1
\PassOptionsToPackage{table,dvipsnames}{xcolor}
\PassOptionsToPackage{dvipsnames}{colortbl}
\documentclass{article}
\usepackage[T1]{fontenc}
\usepackage{iclr2027_conference,times}
\usepackage{amsmath,amsfonts,bm}

\def\eqref#1{equation~\ref{#1}}

\def\1{\bm{1}}

\DeclareMathAlphabet{\mathsfit}{\encodingdefault}{\sfdefault}{m}{sl}
\SetMathAlphabet{\mathsfit}{bold}{\encodingdefault}{\sfdefault}{bx}{n}

\usepackage{hyperref}
\usepackage{url}
\usepackage{booktabs}
\usepackage{tabularx}
\usepackage{arydshln}
\usepackage{amsmath}
\usepackage{amssymb}
\usepackage{enumitem}
\usepackage{amsthm}
\newtheorem{remark}{Remark}
\newtheorem{proposition}{Proposition}
\usepackage{graphicx}
\usepackage{xcolor}
\usepackage{multirow}
\usepackage{xspace}
\usepackage{tikz}
\usetikzlibrary{arrows.meta,positioning}
\usepackage{float}
\usepackage{wrapfig}
\floatstyle{ruled}
\newfloat{algorithm}{t}{loa}
\floatname{algorithm}{Algorithm}
\floatstyle{plain}
\newcounter{algline}


\definecolor{darkcerulean}{rgb}{0.03, 0.27, 0.49}
\definecolor{burgundy}{rgb}{0.7, 0.0, 0.13}
\definecolor{majorelleblue}{rgb}{0.38, 0.31, 0.86}
\definecolor{forestgreen}{rgb}{0.13, 0.55, 0.13}
\definecolor{arylideyellow}{rgb}{0.89, 0.61, 0.06}
\definecolor{arsenic}{rgb}{0.23, 0.27, 0.29}
\hypersetup{
  colorlinks=true,
  citecolor=darkcerulean,
  linkcolor=burgundy,
  urlcolor=majorelleblue
}
\newcommand{\LinkTable}[1]{\hyperref[#1]{Table~\ref*{#1}}}
\newcommand{\LinkFigure}[1]{\hyperref[#1]{Figure~\ref*{#1}}}
\newcommand{\LinkSection}[1]{\hyperref[#1]{Section~\ref*{#1}}}
\newcommand{\LinkAppendix}[1]{\hyperref[#1]{Appendix~\ref*{#1}}}
\newcommand{\LinkEquation}[1]{\hyperref[#1]{Equation~\ref*{#1}}}
\newcommand{\LinkProposition}[1]{\hyperref[#1]{Proposition~\ref*{#1}}}
\newcommand{\LinkRemark}[1]{\hyperref[#1]{Remark~\ref*{#1}}}
\newcommand{\LinkAlgorithm}[1]{\hyperref[#1]{Algorithm~\ref*{#1}}}

\definecolor{TableHeader}{HTML}{F2F2F4}
\definecolor{TableFocus}{HTML}{EFEAF6}
\definecolor{TableInk}{HTML}{3D3A50}
\newcolumntype{L}{>{\raggedright\arraybackslash}X}
\newcolumntype{R}{>{\columncolor{TableFocus}}r}
\newcommand{\tablestyle}{%
  \small
  \renewcommand{\arraystretch}{1.16}%
  \setlength{\tabcolsep}{4pt}%
  \setlength{\aboverulesep}{0pt}%
  \setlength{\belowrulesep}{0pt}%
  \setlength{\extrarowheight}{2pt}%
}

\newcommand{\compacttablestyle}{%
  \footnotesize
  \renewcommand{\arraystretch}{1.08}%
  \setlength{\tabcolsep}{3.5pt}%
  \setlength{\aboverulesep}{0pt}%
  \setlength{\belowrulesep}{0pt}%
  \setlength{\extrarowheight}{1pt}%
}
\newcommand{\tablegroup}[2]{%
  \rowcolor{TableHeader}\multicolumn{#1}{c}{\color{TableInk}\bfseries #2}\\}
\newcommand{\tablehead}[1]{\textbf{#1}}

\newcommand{\method}{\textsc{SWM}\xspace}
\newcommand{\ssg}{System-Surface Graph\xspace}

\title{Software World Models: From Consequence Prediction to Decision Value}

\author{%
Tongli Su$^{1,*}$ \quad Yuntong Hu$^{1,*}$ \quad Liang Zhao$^{1,2,\dagger}$ \\
\textbf{Bowen Zhu$^{2}$ \quad JayaSai Somasundaram$^{2}$ \quad Hasibul Haque$^{2}$} \\
\\
$^{1}$Emory University \quad $^{2}$CausalDynamics.com
}

\iclrfinalcopy

\begin{document}
\maketitle
\lhead{Preprint}
{\renewcommand{\thefootnote}{}%
\footnotetext{\hspace{-1.8em}$^{*}$Equal contribution. $^{\dagger}$Corresponding author: \texttt{liang.zhao@emory.edu}.}}

\begin{abstract}
A coding agent may safely modify one repository while silently breaking
downstream services, libraries, or datastores that depend on it. Exhaustively
running integration tests after every agent action is impractical, so the agent
must predict these failures before executing them. Existing software world
models predict the agent's own observations, while static change-impact
analysis only identifies where a change may propagate. We instead introduce
the \emph{Software World Model} (SWM), which models the broader system affected
by a code change and predicts its \emph{blast set}: the components that the
change will break.
SWM follows three stages: explore, learn, and act. \emph{Explore} executes candidate
changes from restored system states, prioritizing regions where observed
failures contradict the dependency graph. \emph{Learn} fine-tunes a language
model on these execution outcomes to predict downstream breakage.
\emph{Act} converts sampled predictions into per-consumer break probabilities
for change ranking, proactive migration, and deciding when another execution is
worth its cost.
On held-out synthetic systems, SWM improves blast-set F1 from $0.431$ for
static reachability to $0.571\pm0.037$, more than halves ranking regret, and
improves migration return at all nine evaluation checkpoints. The two methods
are complementary: reachability is stronger on dependencies represented in the
graph, while SWM recovers failures caused by couplings the graph misses.
Experiments on held-out real libraries further show that predicting structured
failure outcomes, rather than only scalar risk, is important for downstream
decision quality.

\end{abstract}

\section{Introduction}
\label{sec:intro}

A real software system comprises interdependent repositories, services,
libraries, and datastores. These \emph{components} are coupled through APIs,
schemas, shared data, and configuration. A coding agent, however, typically works inside one
repository and verifies its changes with that repository's
tests~\citep{jimenez2024swebench,yang2024swe,wang2025openhands}. A change can
therefore be locally correct while silently breaking a downstream
\emph{consumer}, a component that depends on the changed interface. Every
local check can pass while that consumer fails when the system runs. Before landing a change, the agent needs to anticipate its
\emph{blast set}, the components that the change breaks.

\begin{figure}[t]
\centering
\includegraphics[width=\linewidth]{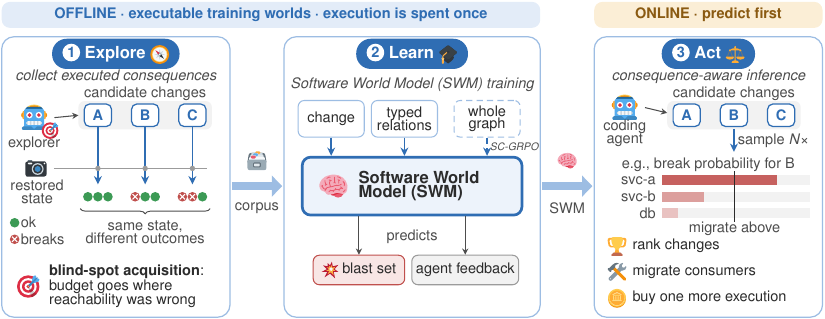}
\caption{\textbf{\method{} closes the loop from exploring executable software
worlds, to learning their change-conditioned consequences, to improving
the coding agent that acts in them.}}
\label{fig:loop}
\end{figure}

Executing every potentially affected component would reveal the blast set, but
does not scale~\citep{memon2017taming,machalica2019predictive}. Classical
change-impact analysis instead identifies what a change may affect from an
extracted dependency graph~\citep{bohner1996impact,rothermel1997safe}, but not
whether a particular consumer actually breaks. Our static baseline, typed reachability (reachability below), marks consumers
connected to the changed interface by dependency paths. It cannot distinguish
a consumer that tolerates a change from one that does not, and misses failures
through couplings the extractor did not recover. We call couplings missing from the extracted graph \emph{hidden channels}.
A \emph{silent break} is a downstream failure missed by repository tests. More fundamentally, any predictor computed from the extracted graph
and change alone carries an error floor given by variation in outcomes among
systems sharing that observation (\LinkRemark{rem:irreducible}).

Existing software world models address the complementary problem: they learn
what the agent itself will observe. CWM learns from execution traces and agent
interactions~\citep{copet2025cwm}, SWE-World predicts execution and test
feedback~\citep{sun2026swe}, and Qwen-AgentWorld predicts tool-call observations
across agent domains~\citep{qwenagentworld}. Their state is the agent-visible
environment, their action is a tool invocation, and their target is the returned
observation. For a locally consistent change, however, that observation can be
correct while a downstream service outside the workspace breaks. We instead
model the \emph{software system the agent changes}, rather than only the
environment it observes.

This shift raises three coupled challenges. \emph{Experience}: downstream
consequences are revealed only by executing affected components, and ordinary
agent trajectories observe only one change from each state.
\emph{Representation}: consequences depend on both what the change does and how
other components use the changed interface, under incomplete structural
observability. \emph{Decision use}: the learned knowledge must improve the agent's
decision before the costly execution that would otherwise reveal the failure.

We introduce the \emph{Software World Model} (\method{}), a learned
consequence predictor, and an \emph{Explore, Learn, Act} framework to build
and use it
(\LinkFigure{fig:loop}). \emph{Explore} executes counterfactual changes from
restored software states to collect their downstream consequences, concentrating
limited execution on informative interventions. \emph{Learn} trains the coding
model on these executed consequences, using the proposed change and typed
software relations to learn change-conditioned system dynamics.
\emph{Act} returns this learned world knowledge to the same coding agent before
future changes, allowing it to anticipate affected consumers rather than
discover them only after execution. The learned consequences are reused for later decisions
(\LinkSection{sec:method}).

Different exploration policies produce different training experience, with
targeted exploration most useful at small budgets. Training on executed
consequences improves the same coding agent under a fixed inference interface: it identifies more affected consumers and ships fewer downstream
failures on held-out software worlds. Further analysis shows that explicit
structure and learned dynamics are complementary. Reachability is strongest on
recorded dependency channels, while learned experience helps where structure is
incomplete. Downstream decision value also depends on \emph{which} errors
a world model makes, not forecast accuracy alone (\LinkSection{sec:exp}).

Our contributions are:
\begin{itemize}[left=0pt]
\item \textbf{Software world modeling.} We formulate system-level consequence
prediction for coding agents and characterize why partial software structure
alone cannot determine downstream behavior (\LinkRemark{rem:irreducible}).

\item \textbf{Explore, Learn, Act.} We introduce a framework that
actively collects counterfactual software experience, trains \method{} on
change-conditioned consequences, and uses the learned dynamics to improve
the agent's decisions before execution (\LinkSection{sec:method}).

\item \textbf{Agent improvement.} We show that learning from executed
consequences improves the same coding agent on held-out software worlds, and
analyze how exploration, observability, and model errors determine downstream
decision value (\LinkSection{sec:exp}).
\end{itemize}

\section{Problem Formulation}
\label{sec:problem}

\subsection{The software world}

A software world is an executable system with checks and available actions.
Its environment $\mathcal{E}$ evolves a true state $s_t = (G^{\star}_t,
x_t)$. Here $G^{\star}_t$ is the true \ssg, a typed, directed multigraph
$G = (V, E, \rho, \omega)$ with nodes $V$, edges $E$, and two layers. The
system layer holds components and shared resources; the surface layer holds
their interfaces, such as endpoints, symbols, and schemas. The map $\omega$ assigns each
surface node its owner, and $\rho : E \to \mathcal{R}$ labels each edge with
its dependency channel, the kind of coupling it represents
(\LinkAppendix{app:ssg}). The behavioural part is a latent execution state
$x_t : V \to \{\texttt{ok}, \texttt{broken}\}$, revealed only by running
something.

\subsection{Agent, actions and observations}

An agent follows $\pi(a_t\mid h_t)$ with history
$h_t=(o_0,a_0,\ldots,o_t)$ and actions
$\mathcal A=\mathcal A_{\mathrm{int}}\cup\mathcal A_{\mathrm{probe}}\cup\mathcal A_{\mathrm{rep}}$:
repository changes (interventions), execution checks (probes), and repairs.
The agent observes
\begin{equation}
o_t = \big(G_t,\; o^{\mathrm{cons}}_t\big),
\qquad G_t = \mathrm{ex}(s_t),
\label{eq:obs}
\end{equation}
where $G_t$ extracts dependencies across the system but omits component code
and some couplings (the hidden channels), and $o^{\mathrm{cons}}_t$ records
feedback from checks already run. Changes keep the edited repository's code
and tests consistent, so local checks pass by construction; downstream
checks do not run by default. Training worlds can be reset, with each
executed intervention and its checks consuming one unit of budget $b$.
Deployment has no reset: local checks cost nothing extra, while each
additional check costs $c_{\mathrm{chk}}$ (\LinkAppendix{app:implementation}).

\subsection{Software world modeling}
\label{sec:problem:belief}

The problem is to anticipate a candidate change's system-wide consequences
before executing it. Define its \emph{blast set} as the components whose
execution state changes,
\begin{equation}
B(s_t, a_t) \;=\; \{\, v \in V \;:\; x_{t+1}(v) \neq x_t(v)
\;\text{when } a_t \text{ is executed in } s_t \,\}.
\label{eq:blast}
\end{equation}
Given $h_t$ and $a_t$, a software world model estimates
\begin{equation}
\boxed{\; p\big(\, B(s_t, a_t),\; o^{\mathrm{cons}}_{t+1} \;\big|\; h_t,\; a_t \,\big) \;}
\label{eq:problem}
\end{equation}
over the blast set and check feedback. At the \emph{forecast step}, a change
lands; its prediction is made beforehand and scored against all executed
downstream effects. The label $\Delta s_t=B(s_t,a_t)$ denotes this blast
set rather than the full state difference.

\label{sec:problem:objective}
The predictive objective is to assign high likelihood to executed consequences,
\begin{equation}
\max_{p}\;\; \mathbb{E}_{(h_t,\, a_t)\,\sim\,\mathcal{D}_{\mathrm{f}}}
\Big[\, \log p\big(\, B(s_t, a_t),\; o^{\mathrm{cons}}_{t+1} \,\big|\, h_t,\; a_t \,\big) \Big],
\label{eq:objective}
\end{equation}
where $\mathcal D_{\mathrm f}$ contains forecast-step records from the executed
corpus $\mathcal D$. Its practical value is whether these forecasts improve
candidate selection and consumer protection before a change lands;
\LinkAppendix{app:protocol} specifies the evaluation criteria.

\label{sec:problem:gap}
Learning and using this distribution raises three unique challenges:
\begin{enumerate}[leftmargin=*,label=\textbf{C\arabic*.},itemsep=2pt,topsep=3pt]
\item \textbf{Experience.} Executed consequences are costly to acquire.
A limited budget must expose failures missed by reachability and compare
alternative changes from the same state, beyond one observed trajectory.
\item \textbf{Representation.} Predicting breakage requires relating changes to
consumer behavior despite missing couplings. The graph alone leaves an
irreducible error floor (\LinkRemark{rem:irreducible}); learning must use
executed outcomes and represent residual uncertainty.
\item \textbf{Decision use.} Before execution reveals failures, comparable
consumer risks must support change selection, preventive migration, and
decisions about paid checks.
\end{enumerate}

\section{Explore, Learn, Act: The SWM Framework}
\label{sec:method}

To address these challenges, we propose an \emph{Explore, Learn, Act}
framework that builds and uses \method{} (\LinkFigure{fig:loop};
\LinkAlgorithm{alg:swm}). For the experience challenge, \emph{Explore}
executes counterfactual changes from restored training states, concentrating
limited execution on informative downstream consequences
(\LinkSection{sec:method:exp}). For the representation challenge,
\emph{Learn} trains a consequence predictor that combines each change with
typed dependencies and observed health to anticipate system-wide effects
(\LinkSection{sec:method:state}). For decision use, \emph{Act} brings these
learned consequences back to the coding agent to guide candidate selection,
preventive migration, and additional checking before a change is committed
(\LinkSection{sec:method:search}). Explore and Learn run offline; their
executed experience supports later agent decisions. The primary study
uses the trained weights directly as the agent (\LinkSection{sec:exp:setup});
the decision rules below separately test how sampled predictions guide action.

\begin{algorithm}[t]
\caption{The \method framework. Explore and Learn run once, offline; Act runs
online before each change.}
\label{alg:swm}
\small
\textbf{Input:} training worlds, budget $b$, explorer $\pi_{\mathrm{e}}$,
pretrained language model $M$, initial executed corpus $\mathcal{D}$, sample
count $N$, costs $c_{\mathrm{act}}, c_{\mathrm{dmg}}, c_{\mathrm{chk}}$.
\begin{list}{\scriptsize\arabic{algline}:}{\usecounter{algline}%
\setlength{\leftmargin}{1.7em}\setlength{\labelwidth}{1.3em}%
\setlength{\itemsep}{0pt}\setlength{\parsep}{0pt}\setlength{\topsep}{2pt}}
\item[] \textit{Explore} (offline; \LinkSection{sec:method:exp})
\item \textbf{while} the explorer has spent fewer than $b$ executions \textbf{do}
\item \quad choose states whose pending interventions have the highest $\hat{d}$ on $\mathcal{D}$ (\ref{eq:blindspot})
\item \quad at each chosen state $s_t$, $\pi_{\mathrm{e}}$ proposes $A_t$; execute every $a \in A_t$ from the restored $s_t$ and add $\mathcal{B}(s_t)$ to $\mathcal{D}$ (\ref{eq:branch})
\item \textbf{end while}
\item[] \textit{Learn} (offline; \LinkSection{sec:method:state})
\item fine-tune $M$ on $\mathcal{D}$ into $W_\phi$ by supervision or group relative policy optimization (GRPO; \ref{eq:reward})
\item[] \textit{Act} (online, before each change; \LinkSection{sec:method:search})
\item the agent proposes candidate changes $A_t$ in its repository
\item sample $W_\phi$ $N$ times per candidate; select $a^\star_t$ by predicted consequence (\ref{eq:rerank})
\item compute $\beta_u$ for $a^\star_t$ (\ref{eq:belief}); while a check pays (\LinkProposition{prop:rules}), check the consumer with the largest expected saving and condition $\beta$ on the answer
\item migrate each $u$ with $\beta_u > c_{\mathrm{act}}/c_{\mathrm{dmg}}$; land $a^\star_t$
\end{list}
\end{algorithm}

\subsection{Explore: branch from restored states where reachability fails}
\label{sec:method:exp}

Explore answers the experience challenge by building the corpus that Learn
trains on. An explorer $\pi_{\mathrm{e}}$ is a policy that proposes changes
in a training world; we evaluate the executed records it collects within
budget $b$. \LinkSection{sec:exp} compares exploration policies at matched
budgets; \LinkAppendix{app:realworld} describes exploration on real software. The
exploration stage combines two operations. Blind-spot acquisition, which
selects executions where previous outcomes contradicted reachability,
picks the states to branch from, and at each, counterfactual branching has
the explorer propose candidate changes that the environment executes in full.
Counterfactual branching supplies same-state comparisons; the real-library
acquisition comparison appears in \LinkFigure{fig:app:real}.

\paragraph{Counterfactual branching.} Given a state, the explorer proposes
$A_t = \{a^1_t, \dots, a^K_t\}$ spanning distinct roles (the action it
prefers, a plausible alternative, a risky one, a subtly wrong one, a
recovery), and the environment executes \emph{all} of them from the same
restored state,
\begin{equation}
\mathcal{B}(s_t) \;=\;
\big\{ \big(s_t,\; a^k_t,\; \Delta s^k_t,\; o^{\mathrm{cons},k}_{t+1}\big) \big\}_{k=1}^{K}.
\label{eq:branch}
\end{equation}
We call $\mathcal{B}(s_t)$ a \emph{branch set}. Branch sets use the explorer
as a sampler of candidates and make observable the quantity that ranking
compares. Because the environment executes every $a^k_t$ from the same
restored $s_t$, the corpus records, for any two candidates,
\begin{equation}
\big\lvert \Delta s^{i}_t \big\rvert - \big\lvert \Delta s^{j}_t \big\rvert
\;=\;
\big\lvert B(s_t,a^i_t) \big\rvert - \big\lvert B(s_t,a^j_t) \big\rvert
\label{eq:contrast}
\end{equation}
with the state held fixed. No confound between what an action does and the
state a policy happened to reach enters the comparison. A single-trajectory
corpus cannot support \LinkEquation{eq:contrast}: it observes one action per
state, so a difference in outcomes conflates the action with the state.

\paragraph{Blind-spot acquisition.} The structural limit in
\LinkRemark{rem:irreducible} motivates collecting outcomes that the graph
alone does not determine. Let $R(G,a)$ denote typed reachability
(\LinkEquation{eq:reachability}). Writing
$\mathcal{P}$ for the pending interventions, the explorer selects
\begin{equation}
\mathcal{A}^{\star} \in
\arg\max_{\mathcal{A}\subseteq\mathcal{P},\,|\mathcal{A}|\leq b}\;
\sum_{a\in\mathcal{A}} \hat{d}\big(\mathrm{tgt}(a)\big),
\qquad
\hat{d}(v) \;=\; \widehat{\Pr}\Big[\,B(s,a)\neq R(G,a)\;\Big|\;\mathrm{tgt}(a)=v\,\Big],
\label{eq:blindspot}
\end{equation}
where the explorer estimates $\hat{d}$ from the executions already in
$\mathcal{D}$. When Explore branches from the states of the selected
interventions, each branch set spends $K$ units of $b$, so the selection
admits at most $b/K$ states. Components with high $\hat{d}$ are reachability blind spots: past executions
frequently disagreed with the graph-based prediction. The criterion needs no label for the candidates it
selects: past disagreement between execution and reachability serves as a
proxy for where the next label is informative. This rule differs from the
decision-aware exploration policy in the primary experiments, which also
uses model uncertainty and novelty (\LinkAppendix{app:loop-protocol}).

\subsection{Learn: fine-tune a language model on executed consequences}
\label{sec:method:state}

\paragraph{Consequence predictor.} Learn fine-tunes a language model $W_\phi$ on executed outcomes. Given
the history and a candidate change, it predicts the blast set and check feedback,
\begin{equation}
W_\phi : (h_t, a_t) \longmapsto
\big(\widehat{\Delta s_t},\; \hat{o}^{\mathrm{cons}}_{t+1}\big),
\label{eq:wm}
\end{equation}
where hats denote predictions. The forecast-step blast set is the primary
target, and check and repair steps supply auxiliary labels
(\LinkAppendix{app:prompt}). $W_\phi$ predicts one step ahead; recursive
rollout is a separate extension (\LinkAppendix{app:recursive}).

\paragraph{Text input.} The default predictor reads a text representation
$S_t^{\mathrm{sem}}$, formed by serialisation $\tau$ of the typed neighbourhood within $r$
hops of the change (at most $m$ edges, the retrieval budget), together with the action and the
observed health $\eta_t$ (component statuses revealed by earlier checks),
\begin{equation}
S^{\mathrm{sem}}_t = \tau\big(\mathcal{N}_t^{r,m}(a_t), a_t, \eta_t\big).
\label{eq:state}
\end{equation}
Typed relations distinguish calls, imports, reads, and subscriptions, which
reachability reduces to reached or not; they also retain weaker traces of
hidden channels. On real libraries,
adding each consumer's usage of the changed symbol raises hidden-channel F1
from $0.142$ to $0.198$ (\LinkFigure{fig:app:real}a). The input ablation in \LinkFigure{fig:app:forecast}c adds dependency relations
to the same observed graph to measure which information helps prediction. The model emits
one structured object holding the predicted blast set, outcome class (no break, silent break, or a break detected by repository
tests), check feedback, and the dependency channel explaining the failure, which puts candidate actions on a common
scale~\citep{qian2026current}.

\paragraph{Training on executed outcomes.}\label{sec:method:rl} A short
supervised warmup teaches the output format. Its loss combines
language modeling with auxiliary losses on the blast set, the outcome class
and the fraction of healthy services; the decision rules do not use the last
of these outputs (\LinkAppendix{app:impl}). The
model then undergoes group relative policy optimization
(GRPO)~\citep{shao2024deepseekmath} to a reward that a program scores against
the executed transition,
\begin{equation}
R_{\mathrm{WM}}(\hat{y})
= w_f R_{\mathrm{format}}
+ w_s F_1\big(\widehat{\Delta s}, \Delta s\big)
+ w_o \mathbf{1}[\hat{c} = c]
+ w_m \tfrac{1}{2}\big(F_1^{\mathrm{tok}}(\hat{o}, o) + \mathrm{chan}(\hat{y})\big),
\label{eq:reward}
\end{equation}
with format, blast-set, outcome-class and observation/channel terms
(\LinkAppendix{app:scgrpo}); no learned reward model enters.
\LinkFigure{fig:app:forecast}a compares this recipe with matched supervised
training on the same corpus.

\paragraph{Graph-conditioned variant.} This variant conditions the
same language model on the whole extracted graph: a relational graph
convolutional encoder~\citep{schlichtkrull2018rgcn} reads the graph, and a
few learned queries over its node embeddings produce \emph{graph tokens},
continuous embeddings that precede the text input. Structure-coupled GRPO
(SC-GRPO), our extension of GRPO, trains the encoder and the language model
jointly on the reward of \LinkEquation{eq:reward}, with one gradient-norm clip
per module (\LinkAppendix{app:scgrpo}; \LinkAppendix{app:impl}). The
companion decision and mechanism analyses pool this variant with
the default model (nine checkpoints);
\LinkTable{tab:app:seeds} compares the two head-to-head, and
\LinkAppendix{app:scgrpo} isolates the effect of per-module clipping.

\begin{figure}[!t]
\centering
\includegraphics[width=\linewidth]{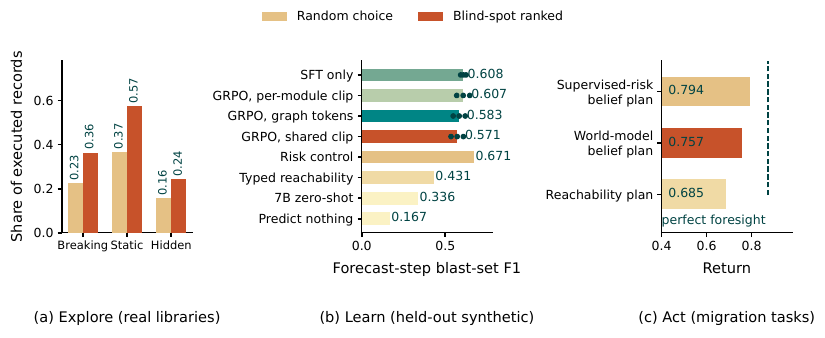}
\caption{Companion evidence across Explore, Learn, and Act.
(a) Fractions of locally passing real-library perturbations that break a
consumer, contradict static prediction, or expose a hidden channel.
(b) Blast-set F1 on 90 held-out synthetic forecasts; dots mark the mean and
mean $\pm$ three-seed half-range.
(c) Explicit-belief planning on 40 migration tasks, with perfect foresight as
a reference. The panels use separate evaluation populations.}
\label{fig:stages}
\end{figure}

\subsection{Act: consult the predictor before a change lands}
\label{sec:method:search}

Before landing a change, the agent proposes candidates $A_t$ and samples $W_\phi$ for each candidate to rank changes, decide
whether to pay for another check, and migrate consumers at risk. A preventive
migration updates a consumer to remain compatible with the pending change;
a repair restores a consumer after it has broken. Only
the checks the agent buys cost execution.

\paragraph{Break probabilities.} Over the consumer set $C_t$, sampling $N$
predicted blast sets estimates each consumer's break probability $\beta_u$.
We call these probabilities the \emph{belief} and use the factorized approximation
\begin{equation}
\hat{p}\big(B \mid h_t,a_t\big)
= \prod_{u \in C_t} \beta_u^{\mathbf{1}[u \in B]}\,(1-\beta_u)^{1-\mathbf{1}[u \in B]},
\qquad
\beta_u = \tfrac{1}{N}\textstyle\sum_{n=1}^{N} \mathbf{1}\big[u \in \widehat{\Delta s}^{(n)}\big],
\label{eq:belief}
\end{equation}
where $\mathbf{1}[\cdot]$ is an indicator. A structural predictor such as $R$ supplies only
$\beta_u \in \{0,1\}$, which is the formal sense in which reachability
cannot express a preference between two candidates it marks alike.

\paragraph{Ranking.} $W_\phi$ imagines each candidate, and the agent lands
only the best,
\begin{equation}
a^{\star}_t \;=\; \arg\max_{a \in A_t} U\big(W_\phi(h_t, a)\big),
\label{eq:rerank}
\end{equation}
with $U$ the negative predicted blast size, or the predicted return for
migration plans.

\paragraph{Preventive migrations and paid checks.} The belief
\LinkEquation{eq:belief} converts preventive action into a decision under
uncertainty rather than a traversal of a set.

\begin{proposition}[One-step decision rules under the belief]
\label{prop:rules}
Let $c_{\mathrm{act}}$ be the cost of migrating a consumer,
$c_{\mathrm{dmg}}$ the cost of leaving it broken, and $c_{\mathrm{chk}}$ the
cost of checking it. Under the factorisation \LinkEquation{eq:belief} the
expected-cost-minimising one-step choices are
\begin{equation}
\text{migrate } u \iff \beta_u > \frac{c_{\mathrm{act}}}{c_{\mathrm{dmg}}},
\qquad
\text{check } u \iff c_{\mathrm{chk}} + \beta_u c_{\mathrm{act}}
< \min\big(c_{\mathrm{act}},\, \beta_u c_{\mathrm{dmg}}\big),
\label{eq:rules}
\end{equation}
and the agent checks the consumer with the largest expected saving until no
check pays.
\end{proposition}

\noindent After each check, the agent conditions the belief on the answer by
reweighting the sampled blast sets that agree with it (\LinkAppendix{app:agent}).

Both rules use the same break probabilities for preventive migration and
paid checks. A set-valued predictor restricts them to $\beta_u\in\{0,1\}$;
probabilistic predictions instead support decisions at the cost thresholds
in \LinkEquation{eq:rules} (\LinkAppendix{app:agent}).

\section{Experiments}
\label{sec:exp}

\subsection{Experimental Setup}
\label{sec:exp:setup}

\paragraph{Models and comparisons.}
All backbones in the main agent comparison have 3B parameters.
We compare the base Qwen2.5-Coder-Instruct agent, its consequence-trained
\method counterpart, and our CWM and Qwen-AgentWorld implementations on
synthetic worlds and held-out real libraries. Baseline adaptation is described
in \LinkAppendix{app:comparators}. The synthetic ablations hold the base
checkpoint, training recipe, and acting interface fixed while varying the
training corpus. Random samples eligible targets; Coverage cycles through
change types; Expert uses true-graph consumer information; Decision-aware
combines uncertainty, disagreement with reachability, and novelty.

\paragraph{Matched synthetic evaluation.}
A training record pairs one executed change with its consequences. Each
360-record corpus has three training seeds. Random and decision-aware
exploration also use 100- and 200-record budgets. These 24 trained arms and
the base checkpoint run on the same 40 tasks from seven held-out worlds,
yielding 1,000 episodes. Each agent first lists candidate consumers for a
fixed change, then selects consumers to migrate. Decoding is greedy, and
execution accounting is identical. The trained checkpoint acts directly,
without external predictions or additional policy fine-tuning.

\paragraph{Metrics and uncertainty.}
Return is $J=1-0.25n_{\mathrm{broken}}-0.05(n_{\mathrm{fixes}}+1)$:
the counts are services left broken and preventive migrations, and the extra
action ships the change. Candidate recall is the fraction of truly affected
consumers in the first call's list. The main table reports means; standard deviations are retained in
\LinkTable{tab:agent-uncertainty}. Synthetic ablations average three seeds within each task and
bootstrap paired return differences by world, using 10,000 resamples.
These intervals apply to the synthetic ablations, not the real-library
comparison. Full task accounting and
seeds appear in \LinkAppendix{app:agent-improvement}; implementation and
metrics are in \LinkAppendix{app:protocol}. Companion prediction and
explicit-belief studies use separate populations and the configuration in
\LinkTable{tab:configuration}.

\subsection{Main Results: Does Learning a Software World Improve the Agent?}
\label{sec:exp:main}
\label{sec:exp:act}

\begin{figure}[t]
\centering
\includegraphics[width=\linewidth]{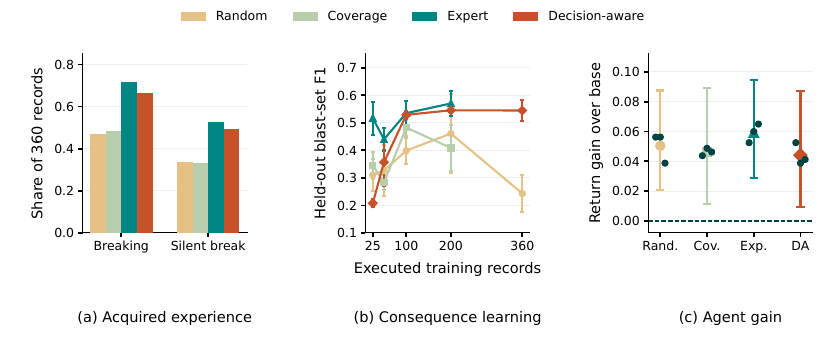}
\caption{Exploration, consequence learning, and agent improvement.
The prediction study in (b) reports three-seed half-ranges;
the agent study in (c) reports paired world-bootstrap 95\% intervals and
individual seed means.}
\label{fig:loop-evidence}
\end{figure}

\paragraph{Exploration determines the experience available for learning.}
\label{sec:exp:explore}
Different policies expose the learner to different software consequences.
Targeted exploration concentrates execution on breaking and locally invisible
transitions (\LinkFigure{fig:loop-evidence}a). These corpus differences also
change prediction at matched budgets: with 100 records, decision-aware
acquisition yields blast-set F1 of $0.529$, versus $0.399$ for random
acquisition. The complete curves show that prediction quality need not grow
monotonically with the record count (\LinkAppendix{app:exploration}).

\begin{wraptable}{r}{0.72\linewidth}
\centering
\caption{Mean agent performance on synthetic and real-library worlds.
Return subscripts give gains over the base agent. Standard deviations appear
in \LinkTable{tab:agent-uncertainty}.}
\label{tab:agent-main}
\compacttablestyle
\scriptsize
\setlength{\tabcolsep}{2pt}
\begin{tabularx}{\linewidth}{@{}Lrrrrrr@{}}
\toprule
\rowcolor{TableHeader}
\tablehead{Agent} & \multicolumn{3}{c}{\tablehead{Synthetic}} & \multicolumn{3}{c}{\tablehead{Real libraries}} \\
\cmidrule(lr){2-4}\cmidrule(l){5-7}
 & \tablehead{Return $\uparrow$} & \tablehead{Recall $\uparrow$} & \tablehead{Broken $\downarrow$}
 & \tablehead{Return $\uparrow$} & \tablehead{Recall $\uparrow$} & \tablehead{Broken $\downarrow$} \\
\midrule
Base agent & $0.721$ & $0.444$ & $0.600$ & $0.453$ & $0.276$ & $0.387$ \\
CWM & ${0.740}_{+0.019}$ & $0.533$ & $0.499$ & ${0.548}_{+0.095}$ & $0.381$ & $0.351$ \\
Qwen-\newline AgentWorld & ${0.754}_{+0.033}$ & $0.559$ & $0.447$ & ${0.571}_{+0.118}$ & $0.446$ & $0.360$ \\
\rowcolor{TableFocus}
\textbf{\method (Ours)} & ${0.773}_{+0.052}$ & $0.586$ & $0.363$ & ${0.613}_{+0.160}$ & $0.499$ & $0.309$ \\
\bottomrule
\end{tabularx}
\par\smallskip
\footnotesize Differences are computed before rounding the displayed means.
CWM~\citep{copet2025cwm} and Qwen-AgentWorld~\citep{qwenagentworld}
are adapted as described in \LinkAppendix{app:comparators}.
\end{wraptable}

\paragraph{Learning executed consequences improves the same coding agent.}
\label{sec:exp:fidelity}
Consequence training raises synthetic mean return from $0.721$ to $0.773$
(\LinkTable{tab:agent-main}). All four full-budget corpora improve return:
each paired interval is positive, as are the point gains for all twelve
training seeds (\LinkAppendix{app:agent-improvement}).

\paragraph{The trained agent anticipates more failures before acting.}
On both datasets, the trained agent identifies more affected consumers and
leaves fewer services broken. Return also charges for migrations, so its gain
reflects the balance between prevention cost and avoided damage. These
measurements support improved anticipation and action without isolating their
individual contributions (\LinkAppendix{app:behavior}).

\paragraph{The same performance pattern holds on real libraries.}
\method has higher mean return and candidate recall, with fewer broken
services, than the base agent and both world-model baselines on each dataset
(\LinkTable{tab:agent-main}). On held-out real libraries, mean return rises
from $0.453$ to $0.613$. This extends the observed benefit beyond procedural systems. Real-library
prediction and exploration diagnostics appear in \LinkAppendix{app:real-results}.

\subsection{Ablation: What Makes Software-World Learning Work?}
\label{sec:exp:ablations}
\label{sec:exp:ablation}

\begin{wraptable}{r}{0.66\linewidth}
\centering
\caption{Consequence training improves the base agent on 40 matched tasks.
Trained rows average three seeds at 360 records; brackets give paired
world-bootstrap 95\% intervals for return differences, computed before rounding.}
\label{tab:loop-main}
\compacttablestyle
\begin{tabularx}{\linewidth}{@{}Lrr@{}}
\hline
\rowcolor{TableHeader}
\tablehead{Training corpus} & \tablehead{Return $\uparrow$} & \tablehead{$\Delta$ return [95\% interval]} \\
\hline
Base agent & 0.721 & $\text{n/a}$ \\
\hdashline
Random & 0.772 & $+0.050\;[+0.021,+0.088]$ \\
\hdashline
Coverage & 0.768 & $+0.046\;[+0.012,+0.089]$ \\
\hdashline
Expert & 0.780 & $+0.059\;[+0.029,+0.095]$ \\
\hdashline
\rowcolor{TableFocus}
Decision-aware & 0.765 & $+0.044\;[+0.010,+0.087]$ \\
\hline
\end{tabularx}
\end{wraptable}

\paragraph{Targeted exploration helps most at the smaller tested budgets.}
Decision-aware acquisition has a larger point gain than random acquisition
at 100 and 200 records. At 100 records, only its interval against the base
agent excludes zero. The gap narrows with more experience, and the ordering
reverses at 360 records (\LinkFigure{fig:loop-budget}). The pattern suggests a sample-efficiency benefit. Intervals compare each
policy with the base agent, rather than with the other policy. Full results
are in \LinkAppendix{app:budget}.

\paragraph{Several exploration policies supply useful training data.}
All four full-budget corpora improve return under the same training recipe
(\LinkTable{tab:loop-main}). Expert exploration gives the largest gain but
uses privileged graph information. Random acquisition also helps, so the
observed benefit is not confined to the targeted acquisition rule. This tests the consequence-training pipeline; an ordinary fine-tuning
control would be needed to isolate its objective.

\paragraph{Richer state helps only when relevant information remains accessible.}
Adding runtime relations lowers learned F1 at the default input budget.
Doubling that budget raises runtime-input F1 from $0.469$ to $0.592$,
above the smaller static input. The pattern is consistent with competition for limited context (\LinkAppendix{app:state-ablation}). Training-recipe comparisons,
including graph tokens and gradient clipping, are reported separately in
\LinkAppendix{app:training-ablation}.

\subsection{Analysis: What Do We Learn About Software World Modeling?}
\label{sec:exp:discussion}
\label{sec:exp:case}

\paragraph{Structure and learned experience resolve complementary uncertainty.}
Reachability has higher F1 on direct propagation, while learned predictions
recover more hidden-channel failures in both evaluated hidden-channel
populations. This contrast supports combining explicit dependencies with executed
experience to predict failures the graph misses
(\LinkAppendix{app:mechanism-analysis}).

\paragraph{Prediction accuracy alone does not determine decision value.}
At matched blast-set F1, learned beliefs produce better migration plans than
a noise-perturbed oracle, but incur higher ranking regret. Pairwise ranking
accuracy also correlates more strongly with regret than aggregate set F1.
The reversal shows why decision evaluation must account for where prediction
errors occur
(\LinkAppendix{app:error-analysis}).

\paragraph{The value of prediction depends on the alternatives to execution.}
The main agent comparison favors \method, but companion studies expose its
boundaries. A supervised risk scorer has higher in-distribution forecast F1
and belief-plan return. Cheap checks of every consumer can also outperform
the learned belief plan. These comparisons use different interfaces from
the main agent evaluation. Their mixed outcomes bound the benefit by the available structure,
execution cost, and decision rule (\LinkAppendix{app:cost-analysis}).

\section{Related Work}
\label{sec:related}

\paragraph{Agent World Models.}
Learned world models let agents train inside or consult imagined
futures~\citep{richens2025general,ha2018world,lecun2022path,hafner2023mastering,hafner2025training,gu2024your,chae2025web}.
Software world models predict execution or tool
feedback~\citep{copet2025cwm,sun2026swe,qwenagentworld,guo2025world,zhang2025agent},
while executable environments supply experience for agent
training~\citep{jimenez2024swebench,yang2025swesmith,swe-gym,badertdinov2025swerebench}.
Our target is the downstream system affected by a change, including failures
outside the agent's workspace. We learn these consequences from execution
and evaluate their value for subsequent decisions.

\paragraph{Software Change Modeling.}
Change-impact analysis and regression-test selection use dependencies to
identify potentially affected
components~\citep{bohner1996impact,rothermel1997safe,memon2017taming,pycg2021,bump2024}.
Graph-conditioned language models provide structural context through graph
representations~\citep{perozzi2024graphtoken,he2024gretriever,hu2024grag,chen2024llaga,li2023blip2}.
SWM uses this structure to predict whether a particular change actually
breaks a consumer. Its evaluation connects to decision-focused prediction,
where quality depends on the decisions a forecast
induces~\citep{elmachtoub2022spo}.

\section{Conclusion}
\label{sec:conclusion}

Coding agents should reason beyond the repository they edit. They should also
anticipate how their changes affect the broader software system. We introduced
Software World Models and an \emph{Explore--Learn--Act} framework for this goal. 
Our experiments show that learning executed consequences improves the same
coding agent under a fixed inference interface. The trained agent identifies
more affected consumers and leaves fewer downstream services broken. We also
find that structure and experience provide complementary information. Explicit
dependencies are reliable when the relevant structure is observed, while
learned dynamics help when that structure is incomplete. Finally, forecast
accuracy alone does not determine decision value. The location and type of
prediction errors also matter.
These results point toward coding agents that do more than react to failures.
By exploring and learning their software worlds, agents can anticipate the
consequences of future changes before acting.

\clearpage

\section*{AI Use Statement}

Generative AI tools were used to assist with code implementation, debugging,
experiment orchestration, and language editing. All generated code was reviewed
by the authors and validated through the project's test and execution pipelines.
AI tools were not used to generate experimental labels or reported results:
software-world consequences were obtained from executed environments, and all
reported numbers were computed from saved experimental outputs. The authors
take full responsibility for the methodology, analysis, claims, and manuscript.

\section*{Ethics Statement}

This work studies coding agents that predict the downstream consequences of
software changes. Our experiments use procedurally generated systems and
permissively licensed open-source software; they involve no human subjects,
personal data, or user-generated content. The proposed techniques may help
identify cross-component failures before software changes are deployed, but
similar capabilities could also reveal structural weaknesses in software
systems. Our experiments are restricted to systems and environments under the
experimenter's control. We encourage deployment only with appropriate
authorization, access controls, and human oversight for consequential software
changes.

\section*{Reproducibility Statement}

Training and evaluation
worlds are disjoint, corpus generation is seeded, and each reported training
condition retains its data split, configuration, random seeds, and model
checkpoint. Executed transitions---rather than model-generated judgments---serve
as supervision for downstream consequences. For the main agent experiments, we
retain per-task outputs and report results across training seeds with
world-clustered uncertainty estimates. The appendices document the software
worlds and graph schema, exploration and evaluation protocols, model training
details, decision accounting, and complete experimental results.

\begingroup
\def\UrlBreaks{\do\/\do\-\do\.\do\_}

\endgroup

\clearpage

\appendix
\section{Complete Evidence for Main Results}
\label{app:loop-protocol}
\label{app:numbers}

The main comparison is reported in \LinkTable{tab:agent-main}. This appendix
separates acquisition, controlled agent improvement, behavioral measurements,
and real-library diagnostics. Companion prediction and explicit-belief
studies use their own populations; \LinkTable{tab:main} summarizes those
results without pooling them with the main agent comparison.

\begin{table}[!htbp]
\centering
\caption{Exploration, forecast, and explicit-decision results. Each row uses its
stated population and comparator.}
\label{tab:main}
\tablestyle
\begin{tabularx}{\textwidth}{@{}lLrrL@{}}
\hline
\rowcolor{TableHeader}
\tablehead{Component} & \tablehead{Evaluation} & \tablehead{Score} &
\tablehead{Comparator or context} & \tablehead{Gain or evidence} \\
\hline
Explore & Real hidden-channel yield & $0.244$ & random: $0.157$ & $+55\%$ \\
\hdashline
Learn & Synthetic blast-set F1 (supervised recipe) & $0.608\pm0.016$ & reachability: $0.431$ & $+0.177$ \\
\hdashline
Learn & Synthetic hidden-channel F1 & $0.304$ & reachability: $0.189$ & $+61\%$ \\
\hdashline
Learn & Real-library hidden-channel F1 & $0.198\pm0.038$ & reachability: $0.137$ & $+45\%$ \\
\hdashline
Act & Candidate-ranking regret $\downarrow$ & $0.593$ & reachability: $1.430$ & $-59\%$ \\
\hdashline
Act & Belief-plan return $\uparrow$ & $0.757$ & reachability: $0.685$ & $+0.072$ \\
\hline
Analysis & Pairwise accuracy and regret correlation & $\rho=-0.93$ & blast-set F1: $-0.56$ & larger correlation magnitude \\
\hline
\end{tabularx}
\end{table}

\subsection{Exploration and consequence learning}
\label{app:exploration}
\paragraph{Acquisition protocol.}
The decision-aware corpus uses the product acquisition rule
\begin{equation}
a^\star=\arg\max_{a\in\mathcal P}
 u_{\mathrm{ep}}(a)\,u_{\mathrm{dec}}(a)\,u_{\mathrm{nov}}(a),
\label{eq:acquisition}
\end{equation}
where $u_{\mathrm{ep}}$, $u_{\mathrm{dec}}$, and $u_{\mathrm{nov}}$ measure
variation across trained models, disagreement with reachability, and lack of
previously executed changes of the same type, respectively.

The decision-aware corpus starts with 60 random records, one per training
world. Each of three rounds scores 960 candidate transitions (16 per world)
and executes the top 100. For each consumer, model uncertainty is the
standard deviation of predicted break probabilities across three model
seeds (16 samples per model); the score averages these deviations.
Reachability disagreement sums the absolute difference between mean break
probability and the binary reachability indicator over consumers predicted to break by any model or marked by reachability. Novelty is $1/(1+n)$, where $n$ counts prior executions with the
same action kind, target kind, and fan-in bucket (the number of incoming observed edges, capped at three). Their product implements
\LinkEquation{eq:acquisition}. The 25- and 50-record prefixes contain only
initial random data. Random, coverage, and expert corpora are separate
exploration baselines; all four full-budget corpora contain 360 records.
Random samples eligible targets without using true consumer information.
Coverage additionally cycles through change types. Expert selects targets
using the true graph, so it is a privileged synthetic-data reference, not a
policy available under partial observability.

\begin{figure}[t]
\centering
\includegraphics[width=\linewidth]{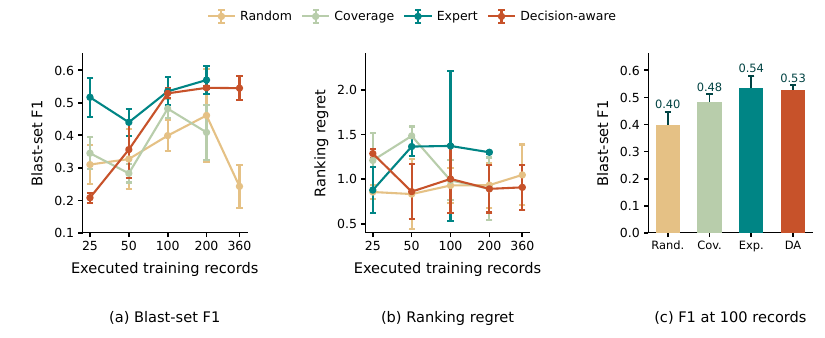}
\caption{Exploration affects prediction and ranking differently.
(a) Blast-set F1 and (b) ranking regret across execution budgets; points are
three-seed means and error bars are half-ranges. (c) The 100-record comparison.
Coverage and expert exploration were not evaluated at 360 records; the first
60 decision-aware records are random. Higher F1 and lower regret are better.}
\label{fig:loop-learning}
\label{tab:loop-learning}
\end{figure}

\paragraph{Companion prediction study.}
SWM-15 and SWM-19 evaluate held-out synthetic forecasts under the
configuration in \LinkTable{tab:configuration}. \LinkFigure{fig:loop-learning} includes every
available exploration/budget cell; absent cells are unevaluated, not interpolated. The random
360-record endpoint uses the matched-recipe rerun in SWM-15.

\subsection{Matched agent improvement}
\label{app:agent-improvement}
The controlled synthetic campaign evaluates 24 trained arms and the same
base checkpoint on 40 tasks. \LinkTable{tab:loop-seeds} retains every training
seed and its paired interval. \LinkAppendix{app:agent-accounting} specifies
aggregation and the world-cluster bootstrap; \LinkAppendix{app:budget}
reports means by corpus and budget.

\begin{table}[p]
\centering
\caption{All 24 trained arms on the same 40 tasks; base-agent return is $0.721$.
Intervals are the source campaign's per-seed paired world-cluster 95\% intervals.}
\label{tab:loop-seeds}
\compacttablestyle
\begin{tabularx}{\linewidth}{@{}Lrrrrr@{}}
\hline
\rowcolor{TableHeader}
\tablehead{Corpus} & \tablehead{Records} & \tablehead{Seed} & \tablehead{Return} & \tablehead{$\Delta$ return} & \tablehead{95\% interval} \\
\hline
Random & 100 & 0 & 0.749 & $+0.028$ & $[-0.009,+0.072]$ \\
Random & 100 & 1 & 0.754 & $+0.033$ & $[+0.005,+0.067]$ \\
Random & 100 & 2 & 0.731 & $+0.010$ & $[-0.023,+0.052]$ \\
Random & 200 & 0 & 0.755 & $+0.034$ & $[+0.011,+0.065]$ \\
Random & 200 & 1 & 0.769 & $+0.048$ & $[+0.018,+0.086]$ \\
Random & 200 & 2 & 0.778 & $+0.056$ & $[+0.025,+0.094]$ \\
Random & 360 & 0 & 0.778 & $+0.056$ & $[+0.025,+0.095]$ \\
Random & 360 & 1 & 0.777 & $+0.056$ & $[+0.025,+0.099]$ \\
Random & 360 & 2 & 0.760 & $+0.039$ & $[+0.011,+0.071]$ \\
\hdashline
Coverage & 360 & 0 & 0.765 & $+0.044$ & $[+0.006,+0.093]$ \\
Coverage & 360 & 1 & 0.770 & $+0.049$ & $[+0.012,+0.088]$ \\
Coverage & 360 & 2 & 0.767 & $+0.046$ & $[+0.007,+0.092]$ \\
\hdashline
Expert & 360 & 0 & 0.774 & $+0.052$ & $[+0.025,+0.082]$ \\
Expert & 360 & 1 & 0.781 & $+0.060$ & $[+0.030,+0.099]$ \\
Expert & 360 & 2 & 0.786 & $+0.065$ & $[+0.032,+0.107]$ \\
\hdashline
Decision-aware & 100 & 0 & 0.768 & $+0.046$ & $[+0.014,+0.084]$ \\
Decision-aware & 100 & 1 & 0.762 & $+0.041$ & $[+0.009,+0.080]$ \\
Decision-aware & 100 & 2 & 0.764 & $+0.043$ & $[+0.009,+0.086]$ \\
Decision-aware & 200 & 0 & 0.768 & $+0.046$ & $[+0.007,+0.094]$ \\
Decision-aware & 200 & 1 & 0.782 & $+0.061$ & $[+0.031,+0.096]$ \\
Decision-aware & 200 & 2 & 0.779 & $+0.058$ & $[+0.029,+0.096]$ \\
Decision-aware & 360 & 0 & 0.774 & $+0.052$ & $[+0.023,+0.091]$ \\
Decision-aware & 360 & 1 & 0.760 & $+0.039$ & $[-0.001,+0.086]$ \\
Decision-aware & 360 & 2 & 0.762 & $+0.041$ & $[+0.005,+0.087]$ \\
\hline
\end{tabularx}
\end{table}

\subsection{Anticipation and post-change failures}
\label{app:behavior}
Candidate recall is measured before preventive migrations. Broken-service
counts measure the outcome after the change lands. All twelve full-budget
seeds improve both measures relative to the base checkpoint, as well as
return. These observations locate behavioral changes but do not decompose
causal contributions of prediction, proposal, and action selection.

\begin{figure}[t]
\centering
\includegraphics[width=\linewidth]{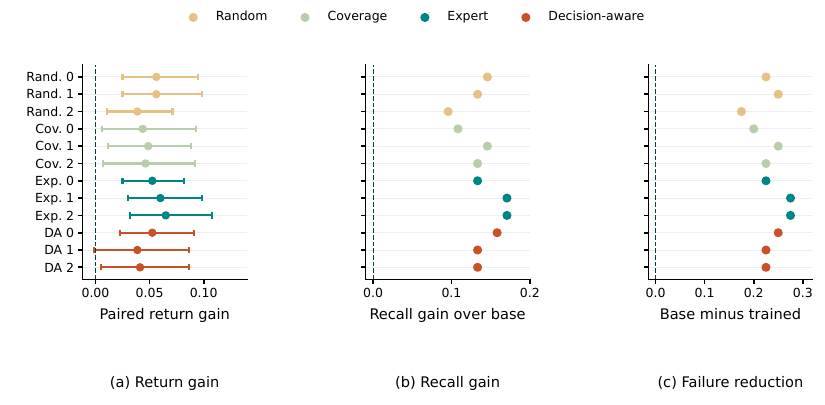}
\caption{All twelve full-budget training seeds improve agent return, candidate recall, and services left broken relative to the base checkpoint.
Panel (a) includes paired world-cluster 95\% intervals; (b--c) show per-seed mean changes on the same tasks.}
\label{fig:loop-controls}
\end{figure}

\subsection{Real-library results}
\label{app:real-results}
\begin{table}[t]
\centering
\caption{Complete agent means and standard deviations for the main comparison.
Values are means, with standard deviations in superscripts where available.
Return subscripts
show the mean difference from the base agent within each dataset. Recall
measures affected-consumer coverage; broken services are counted per task.}
\label{tab:agent-uncertainty}
\compacttablestyle
\setlength{\tabcolsep}{2pt}
\begin{tabularx}{\linewidth}{@{}Lrrrrrr@{}}
\toprule
\rowcolor{TableHeader}
\tablehead{Agent} & \multicolumn{3}{c}{\tablehead{Synthetic worlds}} & \multicolumn{3}{c}{\tablehead{Real libraries}} \\
\cmidrule(lr){2-4}\cmidrule(l){5-7}
 & \tablehead{Return $\uparrow$} & \tablehead{Recall $\uparrow$} & \tablehead{Broken $\downarrow$}
 & \tablehead{Return $\uparrow$} & \tablehead{Recall $\uparrow$} & \tablehead{Broken $\downarrow$} \\
\midrule
Base agent & $0.721$ & $0.444$ & $0.600$ & $0.453$ & $0.276$ & $0.387$ \\
CWM~\citep{copet2025cwm} & ${0.740}_{+0.019}^{\pm0.006}$ & $0.533^{\pm0.007}$ & $0.499^{\pm0.020}$ & ${0.548}_{+0.095}^{\pm0.005}$ & $0.381^{\pm0.005}$ & $0.351^{\pm0.014}$ \\
Qwen-AgentWorld~\citep{qwenagentworld} & ${0.754}_{+0.033}^{\pm0.006}$ & $0.559^{\pm0.022}$ & $0.447^{\pm0.027}$ & ${0.571}_{+0.118}^{\pm0.005}$ & $0.446^{\pm0.018}$ & $0.360^{\pm0.022}$ \\
\rowcolor{TableFocus}
\textbf{\method (Ours)} & ${0.773}_{+0.052}^{\pm0.008}$ & $0.586^{\pm0.017}$ & $0.363^{\pm0.021}$ & ${0.613}_{+0.160}^{\pm0.006}$ & $0.499^{\pm0.015}$ & $0.309^{\pm0.018}$ \\
\bottomrule
\end{tabularx}
\par\smallskip
\footnotesize Differences are computed before rounding the displayed means.
Baseline training is described in \LinkAppendix{app:comparators}.
\end{table}
A set-valued prediction lists all affected consumers in one response;
a per-consumer prediction answers separately for each consumer.
Record-order context retains the input record's ordering, while
change-focused context prioritizes information around the changed interface.
Consumer-usage variants additionally include how the dependent code uses the
changed symbol. Consumer AUROC measures ranking of broken versus unbroken
consumers; the Brier score measures squared error of their break probabilities.

\begin{figure}[t]
\centering
\includegraphics[width=\linewidth]{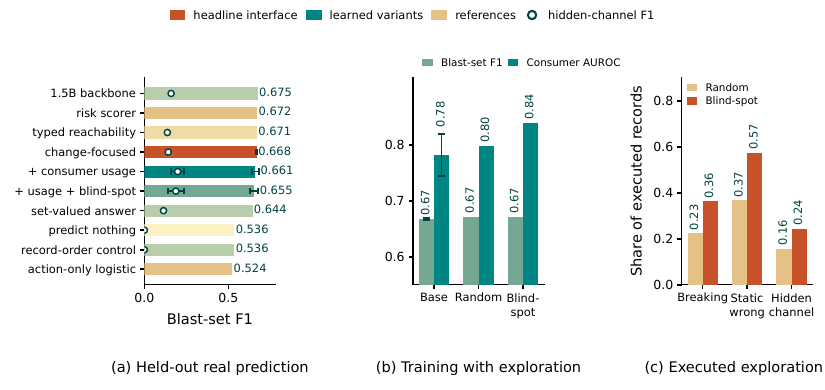}
\caption{Real-library prediction and executed exploration.
(a) Blast-set F1 on 984 held-out records; hollow markers show F1 on the 137
graph-missed breaking records. Error bars retain the reported three-seed
spreads; unmarked learned variants are single runs.
(b) Prediction after adding random or blind-spot exploration, with base-corpus
seed spreads. (c) Acquisition yields from 450 candidate perturbations per
policy on three training worlds, conditional on passing local checks.
Consumer AUROC measures discrimination among broken and unbroken consumers.}
\label{fig:app:real}
\label{tab:baselinesreal}
\label{tab:app:harvested_interface}
\end{figure}

\LinkFigure{fig:app:real} connects prediction to executed exploration.
Consumer-usage context improves hidden-channel F1 even though aggregate F1
changes little. The all-negative control also attains substantial aggregate
F1, making the breaking-record and consumer-level diagnostics necessary for
interpreting the overall score.

\clearpage
\begin{wraptable}{r}{0.64\linewidth}
\centering
\caption{Additional interface diagnostics on the same 984 held-out records.
Columns retain the recorded ``Extracted breaks'' and ``Silent'' measures;
blast-set and hidden-channel F1 appear in \LinkFigure{fig:app:real}a.}
\label{tab:real-interface-details}
\compacttablestyle
\begin{tabularx}{\linewidth}{@{}Lrr@{}}
\toprule
\tablehead{Interface} & \tablehead{Extracted breaks} & \tablehead{Silent} \\
\midrule
Set-valued, record-order context & 0.417 & 0.841 \\
Per-consumer, record-order context & 0.000 & 1.000 \\
Per-consumer, change-focused context & 0.586 & 0.808 \\
\quad + consumer usage & 0.575 & 0.780 \\
\quad + usage and blind-spot round & 0.576 & 0.805 \\
Alternative 1.5B backbone & 0.689 & 0.799 \\
Typed reachability & 0.675 & 0.808 \\
\bottomrule
\end{tabularx}
\end{wraptable}

\paragraph{Exploration yields and probability quality.}
Local checks pass for 337 random and 332 blind-spot perturbations; these are
the denominators for panel (c). Blind-spot selection increases all three
yields, but its downstream benefit is metric-dependent: it gives the highest
consumer AUROC, whereas the random round gives the lowest Brier score.
Brier scores for the base, random-augmented, and blind-spot-augmented corpora
are 0.177, 0.143, and 0.164; their hidden-channel F1 values are
0.142, 0.161, and 0.145. These differences prevent interpreting the AUROC
gain as an improvement in every aspect of probability quality.

\section{Complete Ablations}
\label{app:ablations}

\subsection{Exploration policy and execution budget}
\label{app:budget}
The paired synthetic comparison changes the training corpus and budget
under a fixed agent interface. The three training seeds are averaged within
each task before world-cluster resampling. Prediction learning curves are
in \LinkAppendix{app:exploration}; this subsection reports agent decisions.

\begin{table}[t]
\centering
\caption{All corpus budgets in the agent comparison, averaged over three seeds.
Intervals are paired world-bootstrap 95\% intervals against base-agent return $0.721$.}
\label{tab:loop-budget-all}
\tablestyle
\begin{tabularx}{\linewidth}{@{}Lrrrrr@{}}
\hline
\rowcolor{TableHeader}
\tablehead{Corpus} & \tablehead{Records} & \tablehead{Return} & \tablehead{$\Delta$ return [95\% interval]} & \tablehead{\shortstack[r]{Candidate\\recall}} & \tablehead{\shortstack[r]{Broken\\services}} \\
\hline
Random & 100 & 0.745 & $+0.023\;[-0.008,+0.062]$ & 0.573 & 0.342 \\
\hdashline
Random & 200 & 0.767 & $+0.046\;[+0.018,+0.082]$ & 0.594 & 0.342 \\
\hdashline
Random & 360 & 0.772 & $+0.050\;[+0.021,+0.088]$ & 0.569 & 0.383 \\
\hdashline
Coverage & 360 & 0.768 & $+0.046\;[+0.012,+0.089]$ & 0.573 & 0.375 \\
\hdashline
Expert & 360 & 0.780 & $+0.059\;[+0.029,+0.095]$ & 0.603 & 0.342 \\
\hdashline
\rowcolor{TableFocus}
Decision-aware & 100 & 0.765 & $+0.043\;[+0.013,+0.081]$ & 0.607 & 0.342 \\
\hdashline
\rowcolor{TableFocus}
Decision-aware & 200 & 0.776 & $+0.055\;[+0.023,+0.094]$ & 0.598 & 0.350 \\
\hdashline
\rowcolor{TableFocus}
Decision-aware & 360 & 0.765 & $+0.044\;[+0.010,+0.087]$ & 0.586 & 0.367 \\
\hline
\end{tabularx}
\end{table}

\begin{figure}[t]
\centering
\includegraphics[width=\linewidth]{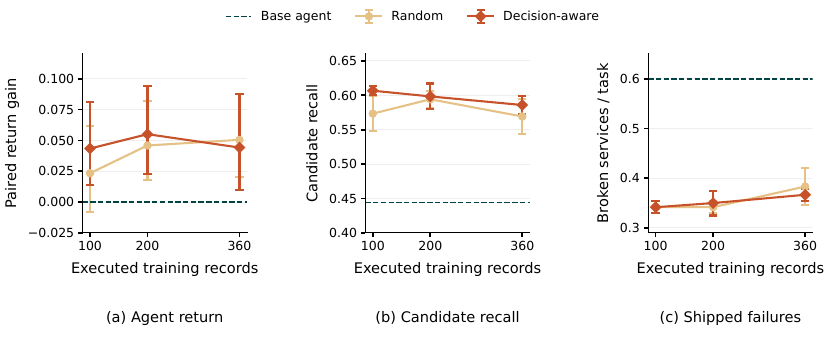}
\caption{Training budget and agent behavior on the same 40 tasks.
Return gains use paired world-bootstrap 95\% intervals; candidate recall
and broken-service counts show three-seed half-ranges, with the base agent as the dashed reference.}
\label{fig:loop-budget}
\end{figure}

\subsection{State information and input budget}
\label{app:state-ablation}
The input configurations in \LinkFigure{fig:app:forecast}c are nested:
\texttt{tree} includes ownership relations; \texttt{imports} adds package
and symbol dependencies; \texttt{calls} adds API calls and schema relations;
\texttt{semantic} includes all extracted static relations; and
\texttt{runtime} adds simulated deployment settings and calls missed by static
extraction.
These are graph-input configurations, not different prediction targets.

\paragraph{Coverage and input budget.}
For the five input configurations in panel (c), mean edge counts are
27.3, 42.2, 46.9, 70.0, and 96.7; traversal precision is
1.000, 1.000, 1.000, 0.778, and 0.803, while recall is
0.167, 0.250, 0.495, 0.574, and 0.691.
The richer runtime input improves traversal coverage but reduces learned F1
at the default retrieval budget; increasing that budget recovers performance.

\subsection{Training recipe and representation}
\label{app:training-ablation}
\label{app:variants}
\begin{figure}[t]
\centering
\includegraphics[width=\linewidth]{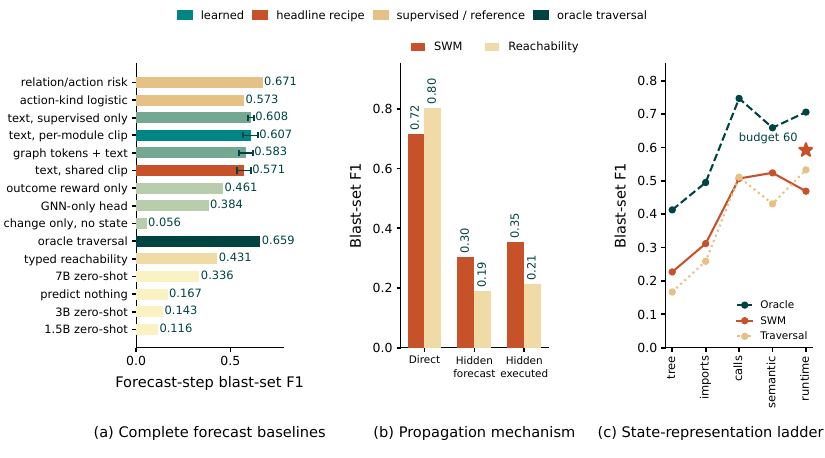}
\caption{Forecast accuracy depends on both propagation mechanism and input information.
(a) Baselines on 90 held-out synthetic forecast steps; error bars show
three-seed half-ranges, and unmarked learned variants are single diagnostic runs.
(b) Direct and hidden-channel F1 across nine checkpoints, with 360, 315,
and 819 seed-record pairs in the three groups, respectively.
(c) Nested graph inputs under identical labels; the star increases the runtime
input budget from 30 to 60. Oracle traversal restores hidden graph channels.}
\label{fig:app:forecast}
\label{tab:baselines}
\label{tab:app:mechanism}
\label{tab:app:ladder}
\end{figure}

\LinkFigure{fig:app:forecast} compares prediction recipes, propagation
mechanisms, and input configurations. Relation/action logistic regression has
the highest aggregate F1 in this companion study. Learned predictors improve
on reachability for hidden channels but trail it on direct propagation,
so their value depends on the mechanism being predicted.

\begin{table}[htbp]
\centering
\caption{Per-module against shared gradient clipping for the companion
recipe, three paired training seeds, forecast-step blast-set F1 on the $90$ held-out
synthetic forecast steps. The graph-token arm uses $K_g{=}8$ learned attention queries
and per-module clipping.}
\label{tab:app:seeds}
\tablestyle
\begin{tabularx}{\linewidth}{L|rrrr}
\hline
\rowcolor{TableHeader}
\tablehead{Configuration} & \tablehead{Seed 0} & \tablehead{Seed 1} & \tablehead{Seed 2} & \tablehead{Mean $\pm$ half-range} \\
\hline
\tablegroup{5}{Learned configurations}
\hdashline
\textbf{text input, per-module clip} & 0.642 & 0.614 & 0.565 & $0.607 \pm 0.039$ \\
\textbf{graph tokens $+$ text input, per-module clip} & 0.621 & 0.549 & 0.579 & $0.583 \pm 0.036$ \\
\rowcolor{TableFocus}
\textbf{text input, shared clip (preregistered)} & 0.565 & 0.537 & 0.611 & $\mathbf{0.571 \pm 0.037}$ \\
\hline
\tablegroup{5}{Structure}
\hdashline
\textbf{typed reachability} & 0.431 & 0.431 & 0.431 & 0.431 \\
\hline
\end{tabularx}
\end{table}

\paragraph{Three-seed consequence learning.}
\LinkTable{tab:app:seeds} reports all seeds for the text-input and graph-token
recipes. Their mean blast-set F1 values range from $0.571$ to $0.607$, above
typed reachability's $0.431$. Seed identifiers denote training repetitions,
not bit-reproducible statistical pairs; the table retains the full variation.
Per-module clipping controls the influence of auxiliary gradient norms and
is an optimizer design, rather than a bound on policy divergence.

\section{Analysis and Failure Modes}
\label{app:analysis}

\subsection{Direct and hidden propagation}
\label{app:mechanism-analysis}
\LinkFigure{fig:app:forecast}b compares the same learned checkpoints with
reachability within each propagation population. Reachability performs
better on direct channels, while learned predictions perform better on
hidden channels. The counts repeat each record across nine checkpoints;
they are not counts of independent worlds.

Across the nine checkpoints, 482 of 565 missed positives (85\%) and
353 of 366 false positives (96\%) lie outside typed reachability.
Even when the blast set is correct, outcome classification is correct on
only 243 of 297 cases (82\%). These diagnostics distinguish missing structural
coverage from errors in predicting consequences.

\subsection{Forecast errors and decision value}
\label{app:error-analysis}
The correlation in \LinkTable{tab:main} is Spearman correlation between
pairwise ranking accuracy and ranking regret across checkpoints and controls;
pairwise accuracy excludes equal-damage pairs and gives half credit to
predicted ties.

\LinkFigure{fig:app:decision}c compares learned predictions with a noise-perturbed
oracle at matched blast-set F1. Ranking and migration depend on where
errors occur, which motivates directly measuring agent decisions in the
primary experiment rather than using blast-set F1 as a substitute.

At matched ranking F1 of 0.525, the perturbed oracle incurs less regret
than the learned checkpoints. In planning, the direction reverses at matched
F1 of 0.546: learned beliefs yield higher return, including relative to an
oracle with F1 0.593. A low-fidelity oracle at F1 0.213 still has ranking
regret 1.343, below reachability's 1.430. The supervised controls change less
under the clean-to-full-noise intervention: ranking regret rises from
0.165 to 0.258, and planning return falls from 0.794 to 0.764.
Thus matching set accuracy does not match the decision consequences of errors.

\clearpage
\subsection{Execution costs and explicit-belief decisions}
\label{app:cost-analysis}
\label{app:agent}

\begin{figure}[t]
\centering
\includegraphics[width=\linewidth]{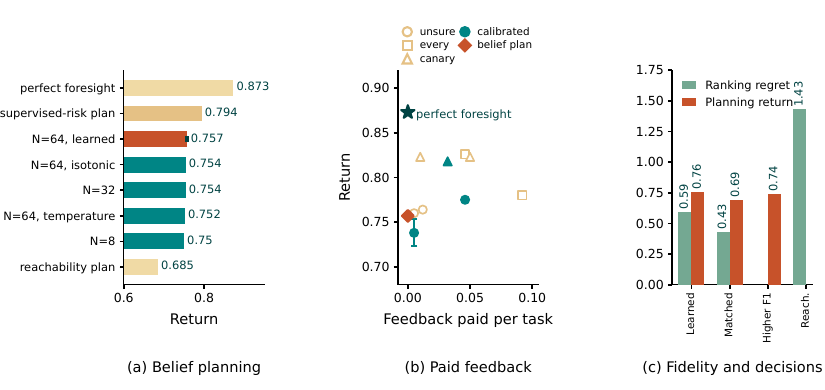}
\caption{Explicit-belief decisions and prediction-error diagnostics.
(a) Mean return for preventive-migration plans on 40 held-out tasks; $N$ is
the number of sampled blast sets. The learned $N=64$ error bar is the
three-seed standard error. (b) Return versus mean expenditure on paid checks:
shape identifies the check strategy and filled markers denote calibrated
beliefs; the displayed error bar is a standard error.
(c) Noise-perturbed oracles matched to learned F1 have lower ranking regret
but lower planning return. Ranking and planning use separate populations;
lower regret and higher return are better.}
\label{fig:app:decision}
\label{tab:feedback}
\label{tab:app:erosion}
\end{figure}

\paragraph{Use consequence probabilities to choose preventive actions.}
We apply the rules in \LinkProposition{prop:rules} to sampled break
probabilities. A paid check reveals one consumer's outcome; a system-wide trial execution
(called a canary in \LinkFigure{fig:app:decision}) reveals the complete blast set. Each observation conditions the sampled belief before the
remaining migrations are chosen.

\paragraph{Measured consequences support better plans than reachability.}
\LinkFigure{fig:app:decision}a--b compares planning and feedback on
40 held-out tasks. Sampled learned beliefs achieve return $0.757$ at
$N=64$, versus reachability's $0.685$. The supervised-risk and perfect-foresight
references are included alongside all calibration and paid-feedback arms,
so additional information is evaluated together with its execution cost.

\paragraph{Evaluation protocol.}
A perfect-foresight plan uses the true blast set before the change.
The learned rows average three training seeds except the marked temperature
scaling run; $\pm$ is their standard error. Calibration adjusts predicted break probabilities using training worlds
only: isotonic calibration fits a monotone mapping, whereas temperature
scaling rescales prediction logits. Costs follow \LinkAppendix{app:protocol};
per-consumer checks cost either 0.01 or 0.02. In that order, uncalibrated
``check where unsure'' gives (return, mean expenditure) of (0.764, 0.012)
and (0.760, 0.005); calibrated checks give (0.775, 0.046) and
($0.738\pm0.015$, 0.005). Checking every consumer gives (0.826, 0.046)
and (0.780, 0.092). Always running a canary gives (0.823, 0.050);
uncalibrated and calibrated belief-triggered canaries give (0.823, 0.010)
and (0.818, 0.032), respectively.

\subsection{Structural limits of consequence prediction}
\label{app:structural-limits}

The typed-reachability baseline follows consumer-dependency paths backwards
from the nodes modified by the change, $\mathrm{tgt}(a_t)$:
\begin{equation}
R(G_t, a_t) = \{u \in V : u \text{ has a consumer-dependency path to } \mathrm{tgt}(a_t)\},
\label{eq:reachability}
\end{equation}
It is a deterministic function of the extracted graph and the change;
\LinkAppendix{app:ssg} defines the dependency relations it follows.

To locate where accuracy beyond reachability must come from, we call a
predictor \emph{structural} if it uses only the extracted graph and the
action, that is, measurable with respect to $(G_t,a_t)$; reachability is
one, and so is any learned model restricted to these inputs.

\begin{remark}[Structural predictors carry irreducible error]
\label{rem:irreducible}
Let $f$ be any predictor with values in $[0,1]^V$ that is measurable with
respect to $(G_t,a_t)$, as the indicator vector of $R$ is. Then $f$ is
constant on states sharing an extracted graph, and for the squared loss
$\ell(B,f)=\lVert\mathbf{1}_B-f\rVert^2$ on the indicator vector of $B$,
whose expectation the conditional mean minimises,
\begin{equation}
\mathbb{E}\big[\ell\big(B(s_t,a_t),\,f(G_t,a_t)\big)\big]
\;\geq\;
\mathbb{E}\big[\operatorname{Var}\big(B(s_t,a_t)\,\big|\,G_t,a_t\big)\big].
\label{eq:irreducible}
\end{equation}
where $\operatorname{Var}$ sums the per-component variances. The right-hand
side is nonzero for two reasons, both properties of the environment rather
than of the estimator. Consumers differ in whether they tolerate a change on a
dependency they share, and the extractor drops channels that carry $34.8\%$
of the broken consumers at our forecast steps (\LinkAppendix{app:corpus}).
\end{remark}

\noindent The bound applies to learned models too: the expected loss is the
floor plus the expected squared distance between $f$ and the conditional mean
of $\mathbf{1}_B$ given $(G_t,a_t)$. No predictor restricted to these inputs
can remove the floor; it can only represent it, as a graded break probability
per consumer. A predictor can narrow the distance, and
reachability leaves it wide by marking every reachable consumer as broken,
tolerant or not, and every consumer reached only through a hidden channel as
safe. Narrowing it requires executed supervision where reachability is wrong;
going below the floor requires information outside the graph, such as executed
outcomes in $h_t$. The evaluated \method also reads the observed health
$\eta_t$, the component health that earlier checks revealed
(\LinkSection{sec:method:state}). When $\eta_t$ carries no further
information about $B$, \method's floor equals the structural one, and \method
can only narrow the distance and represent the floor.

\section{Worlds, Protocol, and Reproducibility}
\label{app:reproducibility}
\label{app:benchmark}

This appendix first defines the software world and its extracted graph. It
then specifies corpus generation, evaluation populations, metrics, and
pre-registration. Implementation and model variants follow in
\LinkAppendix{app:implementation}, decision protocols in
\LinkAppendix{app:agent}, and complete result tables in
\LinkAppendix{app:numbers}.

\subsection{Software worlds}
\label{app:worlds}

We use four complementary families of executable software worlds
(\LinkTable{tab:app:worlds}). The procedural systems provide disjoint training
and evaluation populations with controlled hidden channels. Acme exercises the
same interfaces in a compact multi-repository service system. Online Boutique
provides a polyrepository application built from real services and contracts.
The real-library worlds pair an upstream project with selected downstream
packages that depend on it; their own test suites measure consequences.

\begin{table*}[t]
\centering
\caption{Software worlds used for exploration, training, and evaluation.
Each change must pass the changed repository's own checks before downstream
outcomes are measured.}
\label{tab:app:worlds}
\tablestyle
\begin{tabularx}{\linewidth}{@{}lLLL@{}}
\hline
\rowcolor{TableHeader}
\tablehead{World} & \tablehead{Composition} & \tablehead{Changed interface} &
\tablehead{Executed signal} \\
\hline
Procedural systems & $4$--$7$ services, shared library, datastore, events,
configuration and access control & Nine change families over typed relations &
Repository suites, service health, checks, and true blast sets \\
\hdashline
Acme & Four repositories with three Flask services and one shared library &
API, dependency, database, event, configuration, flag and permission changes &
Local suites plus end-to-end checks \\
\hdashline
Online Boutique & Ten repositories containing nine Go, Node and Python
microservices & gRPC contracts, service wiring, data and latency changes &
Repository tests, service health, direct and web journeys \\
\hdashline
Real-library worlds & Marshmallow with four dependents; OpenTelemetry semantic
conventions with four instrumentation packages & Locally consistent upstream
API and contract changes & Frozen upstream and downstream repository test suites \\
\hline
\end{tabularx}
\end{table*}

\subsection{World construction and evolution}
\label{app:world-evolution}

Each world is frozen at an identified baseline and exposes a declarative action
catalog. An action contains its target surface, materialized patch, expected
local checks, and the checks that can observe downstream effects. Before every
transition, the harness restores the baseline, verifies its health, applies
exactly one selected patch, and reruns the declared checks. The resulting
transition stores the pre-change observation, action identity, patch hash,
local outcome, downstream blast set, check feedback, and restoration
receipt. Repeating a transition starts from the same restored state.

Evolution occurs only through verified transitions. During counterfactual
exploration, several candidates branch from one shared state and only the
selected branch extends the agent history. During sequential exploration, the
next request receives the outcomes of earlier selected actions but never sees
validation or test outcomes. A trained world model may then forecast the
remaining candidates. Its frozen forecasts are returned to the coding agent,
which selects the next action before the harness executes it.

\subsection{System-Surface Graph schema}
\label{app:ssg}

Node kinds and edge relations are tabulated in
\LinkTable{tab:node-kinds} and~\LinkTable{tab:edge-relations}.

\begin{table}[htbp]
\centering
\caption{Node kinds. Every surface-layer node has exactly one owning
system-layer node, which identifies the component that owns each interface.}
\label{tab:node-kinds}
\tablestyle
\begin{tabularx}{\linewidth}{llL}
\hline
\rowcolor{TableHeader}
\tablehead{Layer} & \tablehead{Kind} & \tablehead{Meaning} \\
\hline
\multirow{9}{*}{System}
& \texttt{repo}      & a version-controlled repository \\
& \texttt{service}   & a deployable component \\
& \texttt{library}   & a shared importable component \\
& \texttt{database}  & a datastore instance \\
& \texttt{table}     & a table or collection \\
& \texttt{topic}     & an event stream \\
& \texttt{config}    & a configuration value a component depends on \\
& \texttt{flag}      & a feature flag \\
& \texttt{external}  & a third-party dependency \\
\hline
\multirow{5}{*}{Surface}
& \texttt{endpoint}  & an HTTP/gRPC route a service exposes \\
& \texttt{symbol}    & an exported function or class \\
& \texttt{schema}    & a data contract \\
& \texttt{column}    & a table column \\
& \texttt{scope}     & an authorisation permission \\
\hline
\end{tabularx}
\end{table}

\begin{table}[htbp]
\centering
\caption{Edge relations. The consumer relations are those along which breakage
propagates backwards, and are the ones the reachability baseline traverses.}
\label{tab:edge-relations}
\tablestyle
\begin{tabularx}{\linewidth}{lLLc}
\hline
\rowcolor{TableHeader}
\tablehead{Relation} & \tablehead{src $\rightarrow$ dst} & \tablehead{Recovered from} & \tablehead{Consumer} \\
\hline
\texttt{contains}        & repo $\rightarrow$ service/library & repository layout & \\
\texttt{depends\_on}     & repo/service $\rightarrow$ library  & manifest, lockfile & \checkmark \\
\texttt{imports\_symbol} & service $\rightarrow$ symbol        & import statements & \checkmark \\
\texttt{provides}        & service $\rightarrow$ endpoint      & route declarations & \\
\texttt{calls\_api}      & service $\rightarrow$ endpoint      & client call sites & \checkmark \\
\texttt{publishes}       & service $\rightarrow$ topic         & bus client calls & \\
\texttt{consumes}        & service $\rightarrow$ topic         & bus client calls & \checkmark \\
\texttt{conforms\_to}    & endpoint/topic $\rightarrow$ schema & schema registry & \checkmark \\
\texttt{reads}/\texttt{writes} & service $\rightarrow$ table/column & SQL, migrations & \checkmark \\
\texttt{guarded\_by}     & service $\rightarrow$ flag/config   & configuration references & \checkmark \\
\texttt{requires\_scope} & endpoint $\rightarrow$ scope        & auth middleware & \checkmark \\
\texttt{field\_of}       & column $\rightarrow$ table          & DDL & \\
\hline
\end{tabularx}
\end{table}

\subsection{Corpus generation}
\label{app:corpus}

\paragraph{Systems.} Each system contains $4$--$7$ services and one shared-library
repository with three exported symbols. Services have $2$--$3$ endpoints,
a feature flag, and a timeout configuration. A datastore with two columns
is attached to $75\%$ of services. Synchronous calls form a directed acyclic
graph over the service order, with edge probability $0.35$. Event topics
are published by $40\%$ of services and consumed by $30\%$ of the others.
Cross-service column reads occur with probability $0.18$, modeling the
shared-database antipattern. Half the services have authorisation scopes.

Across the held-out corpus, $34.8\%$ of broken consumer references in the
$90$ forecast-step blast sets are reached only through a dropped channel.

\paragraph{Extractor model.} Channels are dropped from the observed graph at
per-relation rates reflecting real extraction failure modes: \texttt{reads}
$0.60$ (class-level data attributes), \texttt{calls\_api} $0.35$ (attribute-chain
resolution on a client object), \texttt{consumes} $0.30$ (topics bound as wire
literals), \texttt{guarded\_by} and \texttt{requires\_scope} $0.10$,
\texttt{imports\_symbol} $0.05$, \texttt{depends\_on} $0.00$ (manifests parse
reliably). Every dropped edge leaves a weaker dependency trace in the observed graph: a
repository-level \texttt{depends\_on} edge to the provider, plus a base-URL or
topic-name configuration node for API and event channels. These traces can help predict failures without restoring the missing
dependency path (\LinkSection{sec:problem:gap}).

\paragraph{Propagation.} Breakage is computed deterministically on the true
graph, according to the change type. field renames break callers whose recorded
\texttt{fields\_read} includes the renamed field. Symbol renames and breaking
dependency upgrades break importers. Column renames break cross-service
readers. Event-key renames break consumers. Route moves and new authorisation
requirements break callers. Flag flips break callers of the owning service with
probability $0.5$ through a \emph{behavioural} channel with no edge; timeout
tightenings break callers whose fan-out is at least $2$ through a
\emph{latency} channel, likewise edgeless. Breakage cascades one further hop
with probability $0.40$. A broken consumer's own suite catches it with
probability $0.12$, which is what makes most breaks silent.

\paragraph{Rollouts.} Executed-path rollouts have four steps: apply the change
(the forecast step, whose target is the latent blast set), run the repository
suites, run the end-to-end checks, repair one consumer. Branched corpora
execute $K$ explorer-proposed candidates from the same state instead, labelled
\texttt{chosen}, \texttt{alternative}, \texttt{risky}, \texttt{subtly\_wrong},
\texttt{recovery}.

\subsection{Real-world corpora}
\label{app:realworld}

\paragraph{Sources and construction.}
The real-application families in \LinkTable{tab:app:worlds} follow the
restore, execute, and verify procedure in \LinkAppendix{app:world-evolution}.
An upstream change must pass its own test suite before the frozen downstream
suites measure its effects.

\paragraph{Real-library corpus and exploration.}
The real-library study trains on three training worlds built from upstream
projects and their dependent packages and
evaluates prediction on $984$ held-out records drawn from libraries disjoint
from those worlds. Within the $984$ records, $137$ breaking records are
missed by the extracted graph, which gives a real-data hidden-channel
population. Executed exploration follows the acquisition protocol of the
synthetic study: each policy scores $450$ candidate perturbations per round,
and the executed round is added to the training corpus
(\LinkFigure{fig:app:real}). Because the downstream suites are
the projects' own tests, a break recorded here is a real test failure rather
than a simulated propagation event.

\paragraph{Relation to the synthetic population.}
The synthetic agent comparison uses procedural systems with controlled
propagation and inexpensive counterfactual execution. The real-library
comparison in \LinkTable{tab:agent-main} extends the agent evaluation.
Companion prediction and exploration studies use organic dependencies and
repository suites
(\LinkFigure{fig:app:real}a); \LinkAppendix{app:protocol} describes the
feasibility study for agent use.

\subsection{Comparison methods}
\label{app:comparators}

For the main comparison, we trained 3B adaptations of CWM and
Qwen-AgentWorld following the training procedures described in their
papers~\citep{copet2025cwm,qwenagentworld}. These are our implementations at
a common model size: the original CWM has 32B parameters, while
Qwen-AgentWorld releases 35B-A3B and 397B-A17B mixture-of-experts models
(the suffix gives active parameters).

\LinkTable{tab:app:comparators} separates structural, supervised, generative,
and oracle references. All repository-level methods receive the same observed
state and action whenever their input contract permits it.

\begin{table}[htbp]
\centering
\caption{Comparison methods and their roles. Learned controls are fitted only
on training worlds. Oracle rows use information unavailable at decision time.}
\label{tab:app:comparators}
\tablestyle
\begin{tabularx}{\linewidth}{@{}lLLL@{}}
\hline
\rowcolor{TableHeader}
\tablehead{Family} & \tablehead{Method} & \tablehead{Input and output} &
\tablehead{Role} \\
\hline
Structural & Typed reachability & Extracted graph and changed target;
reachable consumers & Reachability baseline \\
\hdashline
Supervised risk & Action-only logistic & Change kind and target count;
consumer probabilities & Action-prior control \\
\hdashline
Supervised risk & Relation/action logistic & Action features, typed degrees,
coverage and hops; consumer probabilities & Strong tabular control \\
\hdashline
Neural structure & Graph neural network (GNN) only & System-Surface Graph; blast set & Graph encoder
without language generation \\
\hdashline
Generative & Zero-shot code models & Serialized state and change; parsed blast
set & Task training against model scale \\
\hdashline
Bounds & Empty, true-graph traversal, and oracle consequence & No failure,
inaccessible structure, or executed outcome & Lower and upper references \\
\hline
\end{tabularx}
\end{table}

\subsection{Evaluation protocol and pre-registration}
\label{app:protocol}

\paragraph{Exploration yields.} A yield is the fraction of executed records
satisfying the stated condition: breaking yield counts changes that break at
least one consumer, and hidden-channel yield counts breaking records missed
by reachability. Static-prediction error counts any mismatch between the
reachability set and the executed blast set. Real-library yields use records
whose upstream checks pass; corpus size and denominators accompany each table.

\paragraph{Blast-set F1.} Each evaluation record carries the change,
the recovered graph, and the blast set observed when the change was
executed in the true system. Blast-set F1 compares the predicted and
executed blast set, computed per record and averaged over records
(record-mean); the companion forecast population is the $90$ step-$0$
forecast steps of the held-out evaluation systems. The direct-channel group contains failures on recovered dependency paths;
the hidden-channel group contains at least one missing dependency channel.
Behavioral/latency records have no hidden channel but depend on response
behavior or timing; no-break records contain no downstream failure. These groups concern graph
coverage, whereas silent/visible labels concern detection by repository tests.

\paragraph{Decision accounting.} A migration episode scores
$\text{return} = 1 - 0.25\,n_{\text{broken}} - 0.05\,(n_{\text{fixes}}+1)
- \text{feedback}$. Here $n_{\text{broken}}$ counts services broken when
the change ships, and $n_{\text{fixes}}$ counts consumers migrated before
shipping. Feedback is the total cost of paid checks at the stated price
(e.g.\ $0.02$ for a consumer check). Belief plans
(\LinkFigure{fig:app:decision}) threshold sampled break probabilities at the
cost ratio and execute the result; hard-selector rows execute a fixed
prediction as the plan.

\paragraph{Ranking.} Each of the $90$ choice sets presents the executed
branches of one counterfactual as candidates. Top-1 is the fraction of
sets whose safest executed candidate is ranked first; pairwise accuracy
is the fraction of candidate pairs ordered correctly; regret is the
expected number of broken services above the best candidate under uniform
tie-averaging.

\paragraph{Agent exploration.} The action catalog and split are frozen before
the first request. At each step the tool-using agent may inspect only eligible
training candidates, selects one action, and receives only that action's
executed downstream observations. Acquisition is measured by verified
transitions, repository test suites executed, breaking yield, hidden-channel yield, and
the utility of the resulting training corpus. Validation and test outcomes are
never returned to the explorer.

\paragraph{Training from acquired experience.} Agent-selected and matched
control corpora use the same optimizer, update count, model initialization,
and evaluation actions. Training seeds measure optimization variation;
exploration seeds measure acquisition variation and are not conflated. A
checkpoint is evaluated only after its training receipt, configuration, corpus
hash, and output parser have been archived.

\paragraph{Companion decision protocols.} Besides the direct agent
evaluation in \LinkAppendix{app:loop-protocol}, we distinguish three protocols.
Ranking chooses among candidates whose outcomes are later executed. Belief
planning converts sampled consequence frequencies into migrations and paid
checks. The real-world agent protocol freezes forecasts before returning them to the
coding agent, records whether the response consulted them, saves the selected
patch before execution, and then verifies patch identity, restoration, and
downstream outcomes. The companion real-world coding-agent protocol tests end-to-end feasibility.
It is distinct from the comparative agent results in \LinkTable{tab:agent-main}.

\paragraph{Prediction-noise diagnostic.} We perturb true blast sets to
compare error patterns at matched blast-set F1. Each prediction-noise axis is swept over eleven noise
levels (\LinkFigure{fig:app:decision}c). The induced blast-set F1 is not evenly
spaced, so a comparison ``at matched blast-set F1'' interpolates the noise-perturbed
oracle's curve at the learned model's own mean F1 rather than reading the
two curves at the same noise level. The same nine checkpoints have mean
blast-set F1 $0.587$ on the $90$ forecast steps, $0.525$ on the $340$ executed
ranking candidates and $0.546$ on the $40$ planning tasks. Both decision
evaluations use this interpolation: the ranking population at F1 $0.525$ and the
planning population at F1 $0.546$. On the
ranking axis a grid level also falls within one step ($0.503$), so that
side does not depend on the interpolation; on the planning axis no level
falls between F1 $0.484$ and $0.603$, so the interpolation is
load-bearing there.

\subsection{Matched agent accounting and training protocol}
\label{app:agent-accounting}
\paragraph{Evidence source and accounting.}
The controlled synthetic campaign is SWM-20, frozen at source revision
\texttt{3fadee6b67ec} of the research repository. In paper revision \texttt{2e32829},
\texttt{evidence/loop\_source.json} stores the 25-arm summary, all 1,000
per-task records, source paths and checksums, companion summaries, and the
training configuration. \texttt{scripts/build\_loop\_evidence.py} verifies
identical task IDs, episode counts, and return accounting, then recomputes
\LinkTable{tab:loop-main} and \LinkTable{tab:loop-budget-all}.
The source summaries round some means to four decimals; paper values are
computed from episodes before rounding. Seven distinct world IDs occur in
the 40-task synthetic sample (a migration task specifies a system state and
one pending change). We sample those seven clusters with replacement
10,000 times (random seed 20260926), retain every task in each sampled
cluster, and recompute the task-weighted paired mean. Training seeds are
averaged within each task before this bootstrap; individual seed estimates
remain visible in \LinkTable{tab:loop-seeds}.

\paragraph{Training and acting protocol.}
All exploration conditions share the update counts, input budgets, adapters,
and reward weights in \LinkTable{tab:configuration}. Low-rank adaptation
(LoRA) uses dropout 0.05. The backbone and two-call acting interface are specified in
\LinkSection{sec:exp:setup}. Candidate proposals in this interface are
consumers, whereas $A_t$ in the ranking protocol contains alternative changes.

\subsection{Implementation scope}
\label{app:implementation}

\paragraph{Training worlds reset; deployment does not.}
\label{sec:problem:regimes}
The reset and execution-cost assumptions are defined in
\LinkSection{sec:problem}. \method targets the regime in which
checking every affected consumer costs more than the breaks that per-consumer
break probabilities let the agent avoid. At the prices of
\LinkAppendix{app:agent} ($c_{\mathrm{chk}}\le 0.02$ against $0.25$ per
broken consumer), checking every consumer still beats migrating by predicted
break probability with the evaluated predictor. We assume that deployment systems and
changes follow the training distribution, and
generalization beyond that distribution is not assumed.

This appendix specifies the consequence-predictor configuration and
graph-conditioned variants; \LinkAppendix{app:loop-protocol} gives the
agent-training protocol.

\subsection{Training configuration and hyperparameters}
\label{app:impl}

\begin{table}[htbp]
\centering
\caption{Consequence-predictor configuration. The default serialises retrieved
state as text; graph-conditioned variants are specified in \LinkAppendix{app:scgrpo}.}
\label{tab:configuration}
\tablestyle
\begin{minipage}[t]{0.485\linewidth}
\vspace{0pt}
\begin{tabularx}{\linewidth}{LL}
\hline
\multicolumn{2}{l}{\textbf{Predictor configuration}} \\
\hline
input & text input (2 hops; $\le 30$ edges) \\
graph mode & none (no graph tokens) \\
base & Qwen2.5-Coder, 0.5B Instruct \\
adapters & LoRA $r{=}16$, $\alpha{=}32$ \\
target modules & q, k, v, o projections \\
precision & bfloat16 \\
prompt budget & 768 tokens \\
answer budget & 192 tokens \\
\hline
\multicolumn{2}{l}{\textbf{Reward weights}} \\
\hline
$w_f$ format & 0.2 \\
$w_s$ blast set & 1.0 \\
$w_o$ outcome & 0.0 \\
$w_m$ observation/channel & 0.0 \\
\hline
\end{tabularx}
\end{minipage}\hfill
\begin{minipage}[t]{0.485\linewidth}
\vspace{0pt}
\begin{tabularx}{\linewidth}{LL}
\hline
\multicolumn{2}{l}{\textbf{GRPO}} \\
\hline
steps & 250 \\
learning rate & $1\times10^{-5}$ \\
group size $G$ & 6 \\
prompts per step & 2 \\
micro-batch size & 1 \\
clip range & 0.2 \\
KL coefficient & 0.02 \\
KL estimator & low-variance $k_3$ \\
loss aggregation & token-mean \\
graph encoder & not used \\
gradient clip & 1.0 shared/global norm \\
\hline
\multicolumn{2}{l}{\textbf{Supervised warmup}} \\
\hline
steps & 200 \\
learning rate & $2\times10^{-4}$ \\
batch size & 4 \\
$\lambda_s$ affected nodes & 1.0 \\
$\lambda_o$ outcome & 0.5 \\
$\lambda_v$ healthy fraction & 0.5 \\
\hline
\end{tabularx}
\end{minipage}
\end{table}

\paragraph{Node features.} Concatenation of a node-kind one-hot ($14$), an
observed-health one-hot ($3$), an action-target indicator ($1$), an action-kind
one-hot ($12$), and a length-$64$ hashed bag of the identifier's alphanumeric
tokens, using a stable hash so features are reproducible across processes.
Identifiers such as \texttt{col:users.users.email} carry real signal about what
a node is and who owns it.

\paragraph{R-GCN update rule.} Starting from input embeddings $h_v^{(0)}$,
the encoder aggregates relation-specific messages~\citep{schlichtkrull2018rgcn}:
\begin{equation*}
\begin{aligned}
u_v^{(l)} &= W_{\mathrm{self}}^{(l)}h_v^{(l)}
+ \frac{1}{d_v}\sum_{r\in\mathcal{R}_{\mathrm{enc}}}
\sum_{u\in\mathcal{N}_v^r}W_r^{(l)}h_u^{(l)},\\
h_v^{(l+1)} &= \mathrm{LayerNorm}\big(h_v^{(l)}+\mathrm{ReLU}(u_v^{(l)})\big).
\end{aligned}
\end{equation*}
Here $\mathcal{R}_{\mathrm{enc}}$ includes forward, reverse and self-loop
relation types; $\mathcal{N}_v^r$ lists incoming neighbours of type $r$,
counting edges, and $d_v=\max(1,\sum_r|\mathcal{N}_v^r|)$ is their total
count. The $W$ matrices are learned transformations; biases and
training-time dropout are omitted. Normalisation is per node, across all
relations, as in the implementation.

\paragraph{Affected-node head.} A per-node multilayer perceptron scores whether
each node is affected,
\begin{equation}
m_v = \mathrm{MLP}_{\psi}\big(h^{(L)}_v\big),
\label{eq:impact}
\end{equation}
where $h^{(L)}_v$ is node $v$'s embedding after $L$ graph layers. Attention over all node embeddings produces graph tokens; this head supplies
an auxiliary prediction of whether each node is affected.

\paragraph{Graph-token scaling.} The projector output is layer-normalised and
scaled to the root-mean-square magnitude of the token embedding matrix. This keeps graph tokens on the scale of the language model's text embeddings during joint training.

These runs use one A100-80GB GPU. Gradient accumulation
provides the effective batch sizes listed in \LinkTable{tab:configuration}.

\subsection{Prompt and output format}
\label{app:prompt}

The user turn contains the retrieved subgraph around the change (nodes
grouped by kind, edges as typed triples, and the observed health), followed
by the action, its target, and a natural-language description of the change. The
system turn fixes the output contract: a single JSON object with fields
\texttt{delta} (the service identifiers that break), \texttt{outcome} (one of
\texttt{no\_break}, \texttt{silent\_break}, \texttt{visible\_break}),
\texttt{observation} (what the checks will report), and \texttt{mechanism}
(which channel carries the breakage and why repository-level checks miss it).
At check steps, \texttt{delta} instead lists newly revealed failures; at
repair steps it lists repaired services. These auxiliary targets describe
check and repair outcomes, not the forecast-step blast set. The blast-prediction metric
uses forecast steps only. Prompts exceeding the budget are truncated from the middle of the edge list, so
the contract at the head and the action at the tail are always preserved.

\subsection{Prediction horizon}
\label{app:recursive}
The evaluated model predicts one transition from the observed state and
refreshes its input after execution. Candidate ranking compares these
one-step forecasts. Recursive latent dynamics are outside the evaluated
model family.

\subsection{Factorization and the SC-GRPO update rule}
\label{app:scgrpo}

The variant of \LinkSection{sec:method:state} couples a graph encoder to the
language model. It factorises the transition,
\begin{equation}
\underbrace{T_t = g_\psi(S^{\mathrm{struct}}_t)}_{\text{graph tokens}}
\qquad
\underbrace{p_\theta\big(\Delta s_t, o^{\mathrm{cons}}_{t+1}
\mid S^{\mathrm{sem}}_t, T_t\big)}_{\text{behavioural consequence}},
\label{eq:factor}
\end{equation}
where $S_t^{\mathrm{struct}}$ encodes the extracted graph with the node
features in \LinkAppendix{app:impl}, and $\phi=(\psi,\theta)$: the graph encoder $g_\psi$ is a four-layer
relational graph convolutional network~\citep{schlichtkrull2018rgcn} with
relation-specific messages and auxiliary outcome and value heads, and the
behavioural factor $p_\theta$ is the same language model as the text-input
model. The graph representation reaches the language model as graph tokens: $K_g$
learned queries $Q\in\mathbb{R}^{K_g\times d}$ cross-attend over the node
embeddings,
\begin{equation}
T_t = P_\psi\big(\mathrm{Attn}(Q, H^{(L)}, H^{(L)})\big),
\qquad
p_\theta(\cdot) = \mathrm{LM}_\theta\big(\cdot \mid [T_t;\ \mathrm{emb}(S^{\mathrm{sem}}_t)]\big),
\label{eq:tokens}
\end{equation}
where $d$ is the node-embedding width, $H^{(L)}$ stacks those embeddings,
and $P_\psi$ projects into the language model's token space~\citep{li2023blip2}. The token budget is fixed and no node is
excluded by a selection step, so the tokens summarise the whole system
while the text input carries the detail near the change.

Vanilla GRPO optimises a language model alone, and the coupled policy is
not one: \LinkEquation{eq:factor} joins a graph encoder to the language model
through the graph tokens, and the encoder carries auxiliary structural losses
of its own during supervised warmup. We therefore train the variant with
\emph{structure-coupled GRPO} (SC-GRPO), which changes the update in the two
places the hybrid policy forces. First, the reward gradient trains both
factors: each micro-batch of the surrogate backpropagates into a detached
copy of the graph tokens $T$, the token gradients are accumulated, and a
single backward pass carries their sum through the encoder, so the
structural factor is shaped by the same verifiable reward as the language
model at the cost of one encoder traversal per step. Second, gradient-norm
clipping is per module during both supervised warmup and GRPO: let $\mathcal{J}$ be the text loss:
negative log-likelihood during supervised warmup, or the negative clipped GRPO
surrogate plus its KL penalty during GRPO. Let
$\mathcal{L}_{\mathrm{aux}}$ be the weighted auxiliary losses during supervised warmup
and zero during GRPO. Writing $\Pi_\gamma$ for clipping a gradient's
Euclidean norm to at most $\gamma$,
\begin{equation}
\begin{aligned}
\nabla_\psi\mathcal{J}
&= \left(\frac{\partial T}{\partial\psi}\right)^{\!\top}
\sum_\mu\nabla_T\mathcal{J}_\mu,\\
\tilde g_\theta &= \Pi_\gamma(\nabla_\theta\mathcal{J}),\qquad
\tilde g_\psi = \Pi_\gamma\big(\nabla_\psi(\mathcal{J}+\mathcal{L}_{\mathrm{aux}})\big).
\end{aligned}
\label{eq:scgrpo}
\end{equation}
Here $T$ is the graph-token tensor, $\mu$ indexes token-weighted micro-batch
losses, and $\tilde g_\theta,\tilde g_\psi$ are the gradients passed to the optimiser.
Clipping applies separately to the parameter groups defined in
\LinkEquation{eq:factor}, rather than to their combined norm. Training-seed results appear in
\LinkTable{tab:app:seeds}; single-run differences need not exceed seed variation.

\end{document}